\documentclass[11pt]{article}

\usepackage[final]{acl}

\usepackage{times}
\usepackage{latexsym}

\usepackage[T1]{fontenc}

\usepackage[utf8]{inputenc}

\usepackage{microtype}

\usepackage{inconsolata}

\usepackage{graphicx}
\usepackage{fontawesome5}
\usepackage{microtype}
\usepackage{hyperref}
\usepackage{url}
\usepackage{booktabs}
\usepackage{amsmath,amssymb}
\usepackage{enumitem}
\usepackage{tcolorbox}
\usepackage{algorithm}
\usepackage{algpseudocode}
\usepackage{graphicx}       
\usepackage{subcaption}   
\usepackage{wrapfig}
\usepackage{verbatim}
\usepackage[table, dvipsnames, svgnames]{xcolor}
\usepackage{bbding}
\usepackage{pifont}
\tcbuselibrary{skins, listings, breakable}
\usepackage{fancyvrb} 
\usepackage{fvextra} 
\usepackage{listings}
\usepackage{titletoc}
\usepackage[toc,page,header]{appendix}
\usepackage{minitoc}
\usepackage{scalerel}
\usepackage{cuted}
\definecolor{darkblue}{rgb}{0, 0, 0.5}
\definecolor{darkgreen}{RGB}{0, 100, 0}
\hypersetup{colorlinks=true, citecolor=darkblue, linkcolor=darkblue, urlcolor=darkblue}

\DefineVerbatimEnvironment{MyBox}{Verbatim}{
frame=single, 
framesep=6pt, 
fontsize=\small, 
breaklines=true, 
breakanywhere=true, 
breaksymbol={}, 
breaksymbolsepleft=0pt, 
breaksymbolsepright=0pt, 
}

\newtcolorbox{case}[1][]{
  colback=black!5,   
  colframe=black!75, 
  boxrule=1pt,       
  arc=4pt,           
  left=6pt, right=6pt, top=6pt, bottom=6pt,
  #1
}

\newcommand{\huggingfaceicon}{\scalebox{1.5}{\scalerel*{\includegraphics{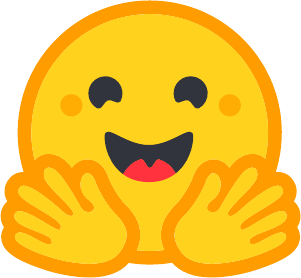}}{X}}}

\definecolor{deepred}{RGB}{180,0,0}
\definecolor{harvestgold}{rgb}{0.85, 0.57, 0.0}

\title{\emph{d}-TreeRPO: Towards More Reliable Policy Optimization for Diffusion Language Models}

\author{\textbf{Leyi Pan}$^{1,2}$\thanks{This work is done during Leyi Pan's internship at Tongyi Lab, Alibaba Group.}, \textbf{Shuchang Tao}$^{2}$, \textbf{Yunpeng Zhai}$^{2}$, \textbf{Zheyu Fu}$^{1}$, \textbf{Liancheng Fang}$^{3}$,\\ \textbf{Minghua He}$^{4}$, \textbf{Lingzhe Zhang}$^{4}$, 
\textbf{Zhaoyang Liu}$^{2}$, \textbf{Bolin Ding}$^{2}$, \textbf{Aiwei Liu}$^{1\dagger}$, \textbf{Lijie Wen}$^{1\dagger}$\\
$^1$Tsinghua University, $^2$Tongyi Lab\includegraphics[height=12pt]{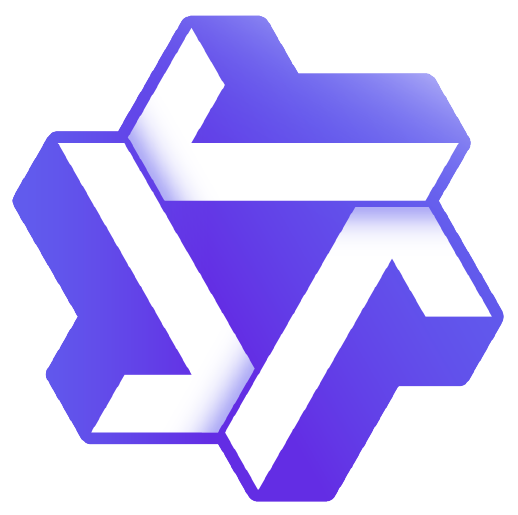}, Alibaba Group, \\$^3$University of Illinois at Chicago, $^4$Peking University\\
\texttt{\small panly24@mails.tsinghua.edu.cn, liuaiwei20@gmail.com, wenlj@tsinghua.edu.cn} \\
\small
\faGithub\; \url{https://github.com/THU-BPM/d-TreeRPO}\quad
\small
\faGithub\; \url{https://github.com/yjyddq/DARE}\\
\small
\huggingfaceicon\; \url{https://huggingface.co/collections/Leyiii/dtreerpo-checkpoints}
}

\begin{document}
\maketitle
{
\let\thefootnote\relax\footnotetext{
$^\dagger$ Corresponding authors. }
}
\begin{abstract}
Reinforcement learning (RL) is pivotal for enhancing the reasoning capabilities of diffusion large language models (dLLMs). However, existing dLLM policy optimization methods suffer from two critical reliability bottlenecks: (1) reward sparsity, arising from coarse or unverifiable signals that impede accurate advantage calculation; and (2) their probability estimates do not account for the gap to the unbiased expectation over all decoding orders, which are intractable to compute. To mitigate these issues, we propose \emph{d}-TreeRPO, a reliable RL framework for dLLMs that leverages tree-structured rollouts and bottom-up advantage computation based on verifiable outcome rewards to provide fine-grained and verifiable step-wise reward signals. Furthermore, we provide a theoretical proof demonstrating that increasing prediction confidence effectively minimizes the gap between unbiased expected prediction probabilities and its single-step forward pass estimate. Guided by this analysis, we introduce a time-scheduled self-distillation loss during training that enhances prediction confidence in later training stages, thereby enabling more accurate probability estimation and better performance. Experiments demonstrate that \emph{d}-TreeRPO outperforms existing baselines and achieves significant improvements across multiple reasoning benchmarks. Specifically, it achieves +86.2\% on Sudoku, +51.6\% on Countdown, +4.5\% on GSM8K, and +5.3\% on Math500 compared to the base model. 
\end{abstract}

\section{Introduction}
\label{sec:intro}

\begin{figure}[h!]
    \centering
    \includegraphics[width=\linewidth]{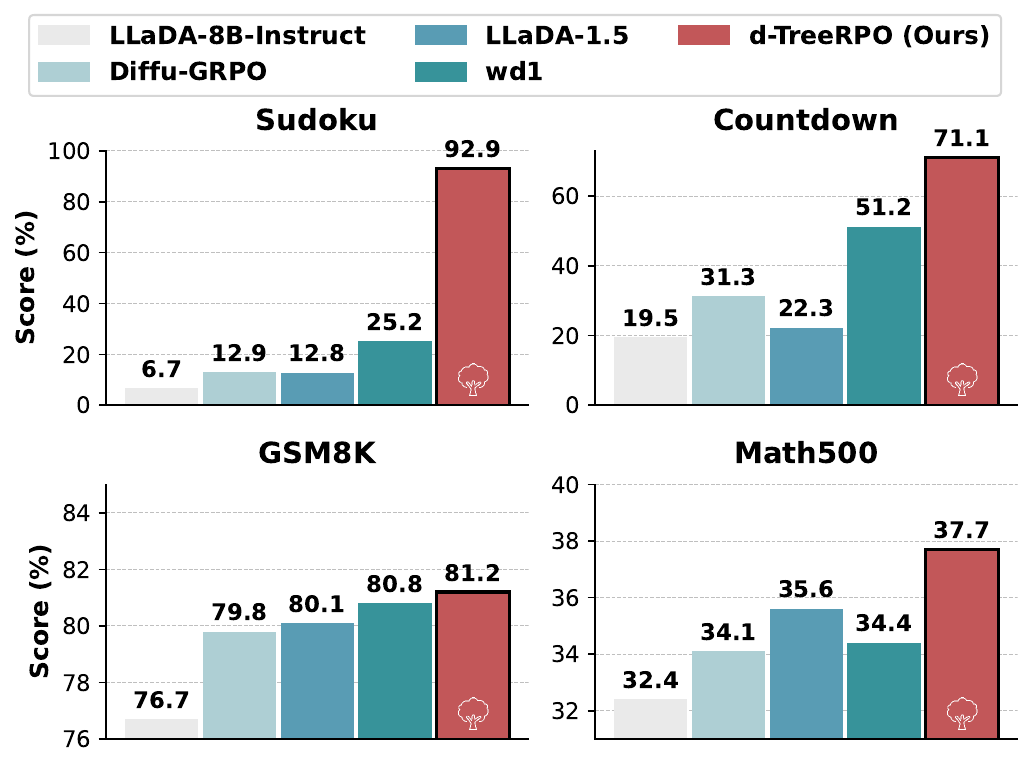}
    \caption{Performance comparison of \emph{d}-TreeRPO with existing dLLM RL methods on four reasoning benchmarks, using LLaDA-8B-Instruct as base model.}
    \label{fig:comparison}
    \vspace{-20pt}
\end{figure}

Diffusion large language models (dLLMs)~\citep{nie2024scaling,ye2023diffusion,zhang2025survey, liu2025wedlm,fang2026locally} provide a parallel decoding alternative to auto-regressive (AR) models. Unlike AR's sequential generation, dLLMs start with masked inputs and iteratively reveal tokens through parallel denoising steps, enabling faster inference. Closed-source models (e.g., Gemini Diffusion, Seed Diffusion~\citep{song2025seed}) achieve 1,400-2,150 tokens/s, while open-source models like LLaDA~\citep{nie2025llada,zhu2025llada1_5,zhu2025lladamoe},  and Dream~\citep{ye2025dream,xie2025dreamcoder} match or exceed AR models with comparable size in performance.

Reinforcement learning (RL) plays a pivotal role in enhancing dLLM reasoning capabilities~\citep{zhao2025d1,huang2025reinforcing,tang2025wd1}. Modern policy optimization frameworks, such as PPO~\citep{schulman2017proximal} and GRPO~\citep{shao2024deepseekmath}, are fundamentally anchored by two core components: the advantage estimate $A(s,a)$ and the action log-probability $\log \pi_\theta(a|s)$. However, both of these components are prone to estimation errors, leading to suboptimal optimization outcomes. To ensure reliable RL for dLLMs, two requirements must be satisfied: \textit{(1) granular and verifiable rewards that ensure accurate advantage calculation}, and \textit{(2) precise estimation of action log-probabilities}.

However, existing RL approaches for dLLMs are hindered by two key reliability bottlenecks that prevent them from meeting the above requirements. \textbf{First, reward design lacks granularity or verifiability}, often relying on sparse outcome rewards~\citep{zhao2025d1,tang2025wd1,wang2025d2,zhu2025llada1_5} or unverified process signals~\citep{huang2025reinforcing,yang2025taming,wang2025revolutionizing}, which can lead to issues like reward hacking. Second, probability estimation remains fundamentally challenging due to dLLMs’ any-order decoding nature. Unlike AR models that decompose sequences via the chain rule for exact probability computation in a single forward pass~\citep{shao2024deepseekmath,zheng2025group,yu2025dapo}, dLLMs require approximations. \textbf{Existing log-probability estimation in dLLMs suffers from unanalyzed approximation errors.} ELBO-based methods~\citep{ou2024your,nie2025llada,yang2025mmada} provide only a biased lower bound and necessitate multi-time forward passes that are often too costly for practical use. While simplified single-step estimators~\citep{zhao2025d1,tang2025wd1} reduce computation, they leave the approximation bias unanalyzed.

\begin{figure*}[t]
    \centering
    \includegraphics[width=\linewidth]{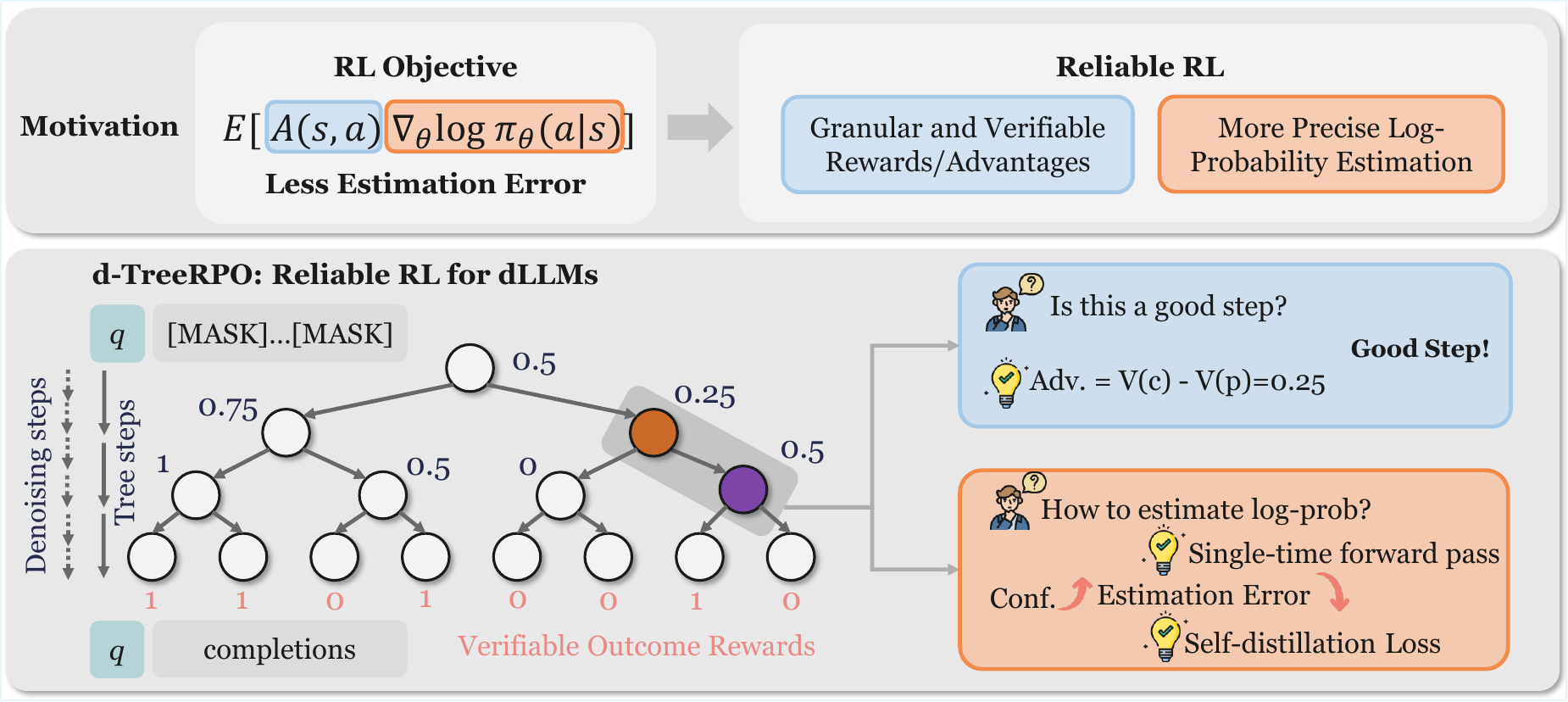}
    \caption{Overview of \emph{d}-TreeRPO. Our framework employs a tree-structured rollout to propagate rewards and compute verifiable step-wise advantages. Guided by theoretical analysis, a time-scheduled self-distillation loss enhances model determinism in later training stages, improving estimation and delivering better performance.}
    \label{fig:method}
    \vspace{-16pt}
\end{figure*}

To address these challenges, we propose \emph{d}-TreeRPO, a reliable policy optimization framework specifically designed for dLLMs. As shown in Figure \ref{fig:method}, we structure rollouts as a tree, with leaf nodes representing terminal states tied to verifiable rewards, and perform bottom-up reward and advantage calculations to obtain granular and verifiable process rewards. For estimating the prediction probability during parent-to-child generation, we theoretically prove that the gap between single-time forward pass probability estimates and the unbiased expectation decreases as the policy becomes more deterministic. However, overly deterministic policies can impair exploration in the early stages of training~\citep{li2025entropy,cui2025entropy,tan2025gtpo}, resulting in a trade-off between exploration and exploitation. To address this, we propose a time-scheduled self-distillation objective that is designed to be weak during the early stages to encourage exploration, and progressively strengthens to promote determinism, thereby improving estimation and overall model performance.

We evaluate \emph{d}-TreeRPO on LLaDA-8B-Instruct~\citep{nie2025llada} and LLaDA-MoE-7BA1B-Instruct~\citep{zhu2025llada} across Sudoku, Countdown, GSM8k~\citep{cobbe2021gsm8k}, and Math500~\citep{lightman2023lets}. Across two model backbones and four reasoning benchmarks, \emph{d}-TreeRPO consistently outperforms baselines. Figure~\ref{fig:comparison} summarizes the results on LLaDA-8B-Instruct, where \emph{d}-TreeRPO achieves +86.2\% (Sudoku), +51.6\% (Countdown), +4.5\% (GSM8k), and +5.3\% (Math500) over the base model. Full experimental results are provided in Section~\ref{sec:experiment}. Moreover, ablation studies further confirm the benefit of the self-distillation loss. In summary, the key contributions are: 
\begin{itemize}
\item We propose \emph{d}-TreeRPO, a more reliable policy optimization algorithm for diffusion language models, which provides fine-grained and verifiable rewards and more reliable probability estimations.
\item We theoretically show that the estimation error of prediction probabilities decreases as the model becomes more confident, and reveal the exploration-exploitation trade-off in dLLM RL. Based on this, we propose a time-scheduled self-distillation loss to relieve the trade-off and improve performance.
\item Extensive experiments and ablations demonstrate significant gains on multiple reasoning benchmarks and verify the effectiveness and practicality of \emph{d}-TreeRPO.
\end{itemize}

\section{Preliminaries}

\textbf{Diffusion Large Language Models.} Diffusion Large Language Models (dLLMs) generate text through an any-order denoising process, gradually reconstructing masked tokens into coherent text until generation is complete. During training, for a corpus sample $y$, a timestep \(u \in [0, 1]\) is randomly sampled. Tokens in $y$ are then masked with probability \(1 - \alpha_u\), where \(\alpha_u\) is a noise schedule (e.g., linear schedule \(\alpha_u = 1 - u\)), resulting in full masking at \(u = 1\) and no masking at \(u = 0\). The corrupted version of the sample is given by:
\begin{equation}
\vspace{-5pt}
\small
    \mathbf{y}_{\text{corrupt},u} = m_u \odot \mathbf{y} + (1 - m_u) \odot \texttt{[MASK]}
\end{equation}
where \(m_u\) is a binary mask vector sampled such that each element \(m_u^{(i)} \sim \text{Bernoulli}(\alpha_u)\), \(\odot\) denotes element-wise multiplication. The model then predicts the original tokens at masked positions by leveraging bidirectional attention. During inference, we discretize denoising timestep into $N$ steps with $\Delta u = 1/N$ and $u_n=1-n\Delta u$ for $n=0,\cdots ,N$. We initialize the completion state at $u_0=1$ as \( z_{u_0} = \{\texttt{[MASK]}\}^L \). For $n=0,\cdots, N-1$, given the current state $z_{u_n}$, the model computes $P_{u_n} = f_{\theta}\big(q,\ z_{u_n})$, samples tokens to reveal (e.g., the top-\( k \) most confident  tokens in masked positions), and updates the state to $z_{u_{n+1}}$, where $u_{n+1} = u_n - \Delta u$. After $N$ steps, we obtain $z_{u_N}$ with $u_N=0$, denotes as $z_0$.


\vspace{2pt}

\noindent\textbf{Group Relative Policy Optimization and its use in dLLMs.} 
GRPO~\citep{shao2024deepseekmath} extends PPO~\citep{schulman2017proximal} with group-relative advantage estimation. The training objective is:
\begin{equation}
\vspace{-9pt}
\small
\begin{split}
&\mathcal{J}(\theta) = \mathbb{E}_{
\substack{
q \sim \mathcal{D}, \\ 
o_i\sim \pi_{\theta_{old}}(\cdot | q)
}
} \Bigg\{\frac{1}{G}\sum_{i=1}^{G}\frac{1}{|o_i|}\sum_{k=1}^{|o_i|} 
\bigg[ \min\Big( r_{i,k}(\theta) A_{i,k},  \\
& \text{clip}\big(r_{i,k}(\theta), 1-\epsilon, 1+\epsilon\big) A_{i,k} \Big) - \beta \mathcal{D}_{\text{KL}}\big(\pi_\theta \| \pi_{ref}\big) \bigg] \Bigg\},
\end{split}
\raisetag{18pt}
\end{equation}
where $\{o_i\}_{i=1}^G$ is a group of $G$ completions drawn from the old policy, group-relative advantage $\small A_{i,k} = R_{i,k} - \text{mean}(\{R_{i,k}\}_{i=1}^G)$, and importance sampling ratio $r_{i,k}(\theta)=\frac{\pi_\theta(o_i^k)}{\pi_{\theta_{old}}(o_i^k)}$. 
Under this training objective and the specifics of dLLMs, two factors are crucial for reliable dLLM RL: \emph{{1) accurate advantage estimation}, including verifiable and fine-grained discrimination}, and \emph{2) precise probability estimation.} In AR models, the prediction probability is straightforward due to fixed decoding order.
In dLLMs, however, the diversity of decoding orders necessitates taking an expectation over all denoising paths. Let $\sigma$ denote a permutation of the unmasking order, sampled from a distribution $\mathcal{Q}(\sigma)$, and let $u_{\sigma,k}$ be the timestep at which position $k$ is unmasked under $\sigma$. Accordingly, the per-token prediction probability is:
\begin{equation}
\vspace{-5pt}
\small
\pi_{\text{dLLM}}(o_i^k|q)
=
\mathbb{E}_{\sigma \sim \mathcal{Q}}
\left[
f_{\theta}^k\!\left(o_i^k\mid q, z_{u_{\sigma,k}}\right)
\right],
\label{eq:unbiased}
\end{equation}
\noindent where $f_\theta^k$ represents the distribution of model's forward function at position $k$, and $z_u$ denotes the completion state at timestep $u$.

\vspace{2pt}

\noindent\textbf{Existing dLLM RL Methods.} While existing works~\citep{zhao2025d1,huang2025reinforcing,tang2025wd1,gong2025diffucoder,rojas2025improving,zhao2025inpainting} attempt to address the aforementioned two aspects, critical limitations remain.

\noindent\emph{1) Reward/advantage estimation:} Many methods directly distribute the final reward uniformly across all tokens.~\citep{zhao2025d1,tang2025wd1,huang2025reinforcing,yang2025taming,xie2025step}. Others introduce more fine-grained supervision via trained reward models~\citep{wang2025revolutionizing}, but may sacrifice verifiability by introducing reward hacking risks. Consequently, existing designs often \textbf{cannot simultaneously provide \emph{fine-grained} and \emph{verifiable} signals}, leading to unreliable advantage estimates.

\noindent\emph{2) Prediction probability estimation:} ELBO-based~\citep{ou2024your} methods use an evidence lower bound as the per-token log-likelihood approximation $\hat{\log\pi_\theta(o^k_i|q)}$ (we use $\hat{\cdot}$ to denote estimated quantities):
\setlength{\abovedisplayskip}{3pt}
\setlength{\belowdisplayskip}{3pt}
\setlength{\abovedisplayshortskip}{3pt}
\setlength{\belowdisplayshortskip}{3pt}
\begin{equation}
\small
\mathbb{E}_{u,z_u}\bigg[\frac{1}{u} \mathbf{1}(z_u^k=\texttt{[MASK]})
\log f_\theta^k(o_i^k| z_u,q)\bigg].
\end{equation}
\noindent Accordingly, the sequence-level ELBO~\citep{rojas2025improving,ou2025principled} is obtained by averaging the token-level estimates. However, ELBO only yields a biased lower-bound and requires computationally expensive multi-round Monte Carlo masking/forward passes for accuracy. For efficiency, methods such as VRPO (LLaDA-1.5)~\citep{zhu2025llada1_5} and BGPO~\citep{lin2025boundary} reduce computation by decreasing the number of Monte Carlo samples or by giving another lower bound of ELBO as an alternative objective. On the other hand, methods such as Diffu-GRPO~\citep{zhao2025d1} and wd1~\citep{tang2025wd1} do per-token approximation using a single forward pass conditioned on the prompt and fully masked completions, which is 
\begin{equation}
\small
\hat{\log\pi_\theta(o^k_i|q)}= \log f_\theta^k(o_i^k|q)
\end{equation}
It greatly reduces computational cost while still achieving noticeable training performance. However, \textbf{none of these methods rigorously quantify the bias between their approximations and the unbiased expected probability} in Equation~(\ref{eq:unbiased}), thereby undermining the reliability of policy-gradient updates.

\vspace{-3pt}
\section{\emph{d}-TreeRPO}

\vspace{-2pt}
\subsection{Tree-Structured RL: Constructing Granular and Verifiable Rewards}
\label{sec:tree-rollout}
\textbf{Tree-Structured Rollout.} To reduce computation during rollout, we introduce a merging parameter \(s\), grouping the \(N\) denoising steps into \(H = N/s\) tree steps. At each tree node (except leaves), \(B\) independent samplings are performed, each involving \(s\) consecutive diffusion steps. This follows the dLLM's inherent sampling mechanism, such as confidence-based sampling in LLaDA~\citep{nie2025llada}. The resulting tree structure produces \(B^H\) leaf nodes for the same prompt, corresponding to \(B^H\) complete generations.

\vspace{2pt}

\noindent\textbf{Rewards and Advantages.} The state value for each leaf node is determined by the verifiable outcome rewards. Based on this, the state value of each tree node is computed bottom-up as the average of its child nodes, i.e.,
\begin{equation}
\label{eq:reward}
\small
V_p = \frac{1}{|C_p|} \sum_{c \in C_p} V_c,
\end{equation}
where \(C_p\) is the set of child nodes of node $p$ and \(V_c\) is the value of child node \(c\). The transition advantage from a parent node to a child node is defined as:  
\begin{equation}
\label{eq:adv}
\small
A_p^c = V_c - V_p = V_c - \frac{1}{|C_p|} \sum_{c' \in C_p} V_{c'},
\end{equation}
where the child nodes under the same parent node are treated as a group, and the relative advantage is calculated within this group.

\vspace{2pt}

\noindent\textbf{Optimization Overview.}
During loss computation, we treat each depth-1 subtree (i.e., a parent node $p$ and its children) as a training unit. 
Specifically, we perform group-wise policy updates over each parent’s child set $C_p$ using a GRPO-style objective. 
Although the update is group-relative at the branch level, the loss is applied token-wise: the parent-to-child transition advantage $A_p^{c}$ is assigned to every newly generated token in the child completion $c$. 
This also requires estimating the probabilities of the newly generated tokens conditioned on the parent state, which we detail in Section~\ref{sec:estimation}.

\vspace{-8pt}
\subsection{Estimation of Conditional Log-Probabilities in Tree Structures}
\vspace{-2pt}
\label{sec:estimation}
\textbf{Single-time Forward Pass Estimation.} For computational efficiency, following Diffu-GRPO~\citep{zhao2025d1} and wd1~\citep{tang2025wd1}, we utilize single-time forward pass to estimate the log-probabilities of the newly generated tokens in child completions given the parent state. Suppose the positions newly decoded in the child node are $d_1, d_2, \dots, d_k$, with the corresponding tokens \(o^{d_1}, o^{d_2}, \dots, o^{d_k}\). Then the estimation is: 
\vspace{2pt}
\begin{equation}
\small
\label{eq:one-step-estimate}
    \hat{\log \pi_\theta(o^{d_i} \mid p)} =\log f_\theta^{d_i}(o^{d_i}\mid p),
\end{equation}
where $\hat{\log \pi_\theta(o^{d_i}|p)}$ denotes the estimation of $\log \pi_\theta(o^{d_i}| p)$, and $f_\theta^{d_i}$ represents the distribution of model's forward function at position $d_i$. $p$ corresponds to the parent node's state, which includes the prompt and the partially masked completion. 



\noindent\textbf{Comparability Issue and Block-wise Adaptation.} 
The design above introduces a comparability issue: when child nodes of the same parent decode different positions within a tree step, the log-probabilities estimated by Equation~(\ref{eq:one-step-estimate}) become incomparable. To address this, we adopt the block-wise decoding approach \citep{arriola2025block} often used in dLLM inference, where tokens within each block are decoded in any order, and blocks are decoded sequentially. By aligning tree-search hyperparameters with the block-wise structure—ensuring the total tree steps \(H\) divide the number of blocks \(L/b\)—each tree step decodes full blocks, restoring log-probability comparability among sibling nodes.

\vspace{-7pt}
\subsection{Relationship Between Prediction Confidence and Estimation Error}
\vspace{-3pt}
\label{sec:theory}

To understand the approximation error of prediction probability for dLLMs, we formally compare the single-time forward pass estimator with the unbiased expected probability in Equation (\ref{eq:unbiased}) under all possible decoding orders.

\vspace{2pt}

\noindent\textbf{Theorem 1} (High-probability Estimation Error Bound).
Let $p$ be the parent state. In one tree step, we decode $k$ masked positions with index set
$\mathcal D=\{d_1,\dots,d_k\}$, yielding realized tokens $o^{d_1},\dots,o^{d_k}$.
A revelation order is a permutation $\sigma=(\sigma_1,\dots,\sigma_k)\sim\mathcal Q$, where $\sigma_j\in\mathcal D$
is the position revealed at sub-step $j$.
Define:
\begin{equation}
\small
q_j(\sigma)
=
f_\theta^{\sigma_j}\!\Big(
o^{\sigma_j}\mid p,\{o^{\sigma_1},\dots,o^{\sigma_{j-1}}\},\texttt{[MASK]}_{\text{else}}
\Big).
\end{equation}
The single-time forward pass estimator is:
\begin{equation}
\small
\hat p_{d}:=f_\theta^{d}(o^{d}\mid p),
\end{equation}
And let $\tau(d,\sigma)$ be the unique index such that $\sigma_{\tau(d,\sigma)}=d$.
For $\delta\in(0,1)$, let $q_{d,1-\delta}$ be the $(1-\delta)$-quantile of $q_{\tau(d,\sigma)}(\sigma)$ under
$\sigma\sim\mathcal Q$, i.e.:
\begin{equation}
    \small
    \Pr_{\sigma\sim\mathcal Q}\!\big(q_{\tau(d,\sigma)}(\sigma)\ge q_{d,1-\delta}\big)\ge 1-\delta.
\end{equation}
And define
\begin{equation}
\small
\epsilon_{d,\delta}:=\max\{1-\hat p_d,\ 1-q_{d,1-\delta}\}.
\end{equation}
Then for $d\in\mathcal D$ with $\epsilon_{d,\delta}\in[0,1)$, we have:
\begin{equation}
\label{eq:theory}
\small
\boxed{
\Pr_{\sigma\sim\mathcal Q}\!\left(
\left|\log\frac{q_{\tau(d,\sigma)}(\sigma)}{\hat p_{d}}\right|
\le
-\log(1-\epsilon_{d,\delta})
\right)\ge 1-\delta.
}
\end{equation}

\vspace{-4pt}

\noindent\textbf{What does this imply?}
With probability at least $1-\delta$ over decoding orders $\sigma\sim\mathcal Q$,
the log-ratio error between the path-wise probability $q_{\tau(d_i,\sigma)}(\sigma)$ and the single-pass estimate
$\hat p_{d_i}$ is controlled within $-\log(1-\epsilon_{d_i,\delta})$. When the model is more \textbf{confident} (i.e., smaller $\epsilon_{d_i,\delta}$),
the right-hand side becomes smaller and \textbf{the bound is tighter}. Since  $p_{d_i}^{\mathrm{true}}=\mathbb E_{\sigma\sim\mathcal Q}[q_{\tau(d_i,\sigma)}(\sigma)]$
is \textbf{an expectation over decoding orders}, this result explains why \textbf{increasing confidence makes single-pass probability estimation more accurate} in practice.
A detailed proof of Formula~(\ref{eq:theory}) is provided in Appendix~\ref{app:proof}.

\vspace{-6pt}

\subsection{Design of Self-distillation Loss}
\vspace{-4pt}
\label{sec:consistency}
\textbf{Exploration-Exploitation Trade-off.} Higher prediction confidence improves probability estimation accuracy (Section~\ref{sec:theory}) but limits exploration~\citep{cui2025entropy,zhang2024entropy}, creating a trade-off in dLLM RL. Early training benefits from greater exploration, while later stages require precise probability estimation for stable convergence and better performance. To balance this, we propose a \textbf{time-scheduled self-distillation loss} that starts with minimal weighting to encourage exploration and gradually increases to focus on high-advantage actions in later training.

\noindent\textbf{Advantage-Weighted Target Distribution.} For a depth-1 subtree, we first select child nodes with positive advantages $C_p^+ = \{c \in C_p \mid A_p^c > 0\}$. The time-dependent temperature $\tau(t)$ controls distribution sharpness:
\begin{equation}
\label{eq:tau-t}
\small
\tau(t) = \tau_{\text{max}} \cdot \left(1 - \frac{t}{T}\right)^\beta, \quad (\beta \in (0,1])
\end{equation}
where $T$ denotes total training steps, and $t$ is the current training step. The advantage-weighted coefficients are computed through:

\begin{equation}
\small
w_c = \frac{\exp\left(A_p^c/\tau(t)\right)}{\sum_{c'\in C_p^+} \exp\left(A_p^{c'}/\tau(t)\right)}, \quad \forall c \in C_p^+
\end{equation}
For each position-token pair $(\sigma_i,v)$ ($\sigma_1,\dots,\sigma_k$ are positions decoded in current tree step, and $v \in \mathcal{V}$, $\mathcal{V}$ is the vocabulary set), we construct the target distribution $P_{\text{target}}^{\sigma_i}(v)$ through advantage-weighted aggregation:
\begin{equation}
\small
P_{\text{target}}^{\sigma_i}(v) = \sum_{c\in C_p^+} w_c \cdot \mathbf{1}[v_c^{\sigma_i} = v], \quad \forall v \in \mathcal{V}.
\end{equation}
\textbf{Time-scheduled Self-distillation Loss.} The final self-distillation loss $\mathcal{L}_\text{distill}$ is then defined as:
\begin{equation}
\label{eq:L-consist}
\small
\lambda(t) \cdot\frac{1}{k}\sum_{i=1}^k D_{\text{KL}} \left( P_{\text{target}}^{\sigma_i}(\cdot) \parallel \pi_\theta^{\sigma_i}(\cdot|p) \right),
\end{equation}
where $k$ is the number of newly generated tokens in this tree step, and $\lambda(t)$ is a time-scheduled weight:
\begin{equation}
\small
    \lambda(t) = \lambda_{\text{max}} \cdot \frac{e^{\gamma t/T}-1}{e^\gamma - 1}.
\label{eq:lambda-1}
\end{equation}
As training progresses, $\tau(t)$ becomes increasingly sharp, favoring selections from child nodes with the highest advantages, while $\lambda(t)$ grows progressively larger, causing the self-distillation loss being increasingly important.


\vspace{2pt}
\noindent\textbf{Complete Loss Function.} Therefore, the overall loss of \emph{d}-TreeRPO is given in Equation~(\ref{eq:d-treerpo-total-loss}). Here, $\mathcal{T}(q)$ denotes the rollout tree for prompt $q$, and $k$ is the number of newly decoded tokens in a tree step.
\begin{figure*}[t] 
\centering
\begin{tcolorbox}[
    width=\linewidth,  
    enhanced,
    colback=cyan!6, 
    colframe=black, 
    arc=1pt,        
    boxrule=1pt,    
    left=0pt, right=0pt, top=-13pt, bottom=0pt 
    ]
\begin{equation}
\label{eq:d-treerpo-total-loss}
\small
\begin{aligned}
\mathcal{L}_{\text{\emph{d}-TreeRPO}}(\theta) = \mathbb{E}_{q \sim \mathcal{D},\;p \in \mathcal{T}(q)} \bigg[-
\frac{1}{|C_p|} \sum_{c \in C_p}\frac{1}{k}\sum_{i=1}^k\Big(\underbrace{ \min \Big( \frac{f_{\theta}^{d_i}(o^{d_i}|p)}{f_{\theta_{old}}^{d_i}(o^{d_i}|p)} A_p^c, 
\operatorname{clip}(\frac{f_{\theta}^{d_i}(o^{d_i}|p)}{f_{\theta_{old}}^{d_i}(o^{d_i}|p)}, 1 - \epsilon, 1 + \epsilon) A_p^c \Big)}_{\text{policy-gradient loss}}
\\
- \underbrace{\beta D_{\text{KL}} \left[ \pi_\theta(\cdot \mid p) || \pi_{\text{ref}}(\cdot \mid p) \right]}_\text{KL loss}\Big)
+ \underbrace{\frac{\lambda(t)}{k} \sum_{i=1}^k D_{\text{KL}} \Big( P_{\text{target}}^{\sigma_i}(\cdot) \parallel \pi_\theta^{\sigma_i}(\cdot|p) \Big)}_{\text{self-distillation loss}}\bigg].
\end{aligned}
\end{equation}
\end{tcolorbox}
\vspace{-10pt}
\end{figure*}

\vspace{-18pt}
\section{Experiments}
\vspace{-4pt}
\label{sec:experiment}
\begin{table*}[t]
  \centering 
  \renewcommand{\arraystretch}{1.2} 
  \resizebox{\textwidth}{!}{%
  \Huge
  \begin{tabular}{l *{8}{c}}
    \toprule
    \textbf{Methods / Datasets} & \multicolumn{2}{c}{\textbf{Sudoku}} & \multicolumn{2}{c}{\textbf{Countdown}} & \multicolumn{2}{c}{\textbf{GSM8k}} & \multicolumn{2}{c}{\textbf{Math500}} \\
    \cmidrule(lr){2-3} \cmidrule(lr){4-5} \cmidrule(lr){6-7} \cmidrule(lr){8-9}
    
    & 256 & 512 & 256 & 512 & 256 & 512 & 256 & 512 \\
    \midrule
    
    LLaDA-8B-Instruct & 6.7 \textcolor{gray}{\huge(+0.0)} & 5.5 \textcolor{gray}{\huge(+0.0)} & 19.5 \textcolor{gray}{\huge(+0.0)} & 16.0 \textcolor{gray}{\huge(+0.0)} & 76.7 \textcolor{gray}{\huge(+0.0)} & 78.2 \textcolor{gray}{\huge(+0.0)} & 32.4 \textcolor{gray}{\huge(+0.0)} & 36.2 \textcolor{gray}{\huge(+0.0)}\\
    \quad + \textit{Diffu-GRPO}~\citep{zhao2025d1}  & 12.9 & 11.2 & 31.3 & 37.1 & 79.8 & 81.9 & 34.1 &  \underline{39.0} \\
    \quad + \textit{VRPO (LLaDA-1.5)}~\citep{zhu2025llada1_5}  & 12.8 & 9.6 & 22.3 & 18.0 & 80.1 & 81.5 & \underline{35.6} & 34.8 \\
    \quad + \textit{wd1}~\citep{tang2025wd1}  & 25.2 & 24.2 & 51.2 & 46.1 & 80.8 & 82.3 & 34.4 & \underline{39.0} \\
    \quad + \textit{SAPO}~\citep{xie2025step}  & 20.3 & 16.1 & 52.0 & 56.3 & 80.6 & 82.1 & 33.8 & 38.4 \\
    \quad + \textit{GDPO}~\citep{rojas2025improving} & \underline{25.7} & 24.8 & \underline{64.1} & \underline{60.2} & \underline{81.1} & 82.0 & \underline{37.0} & 38.5 \\
    \quad + \textit{TraceRL}~\citep{wang2025revolutionizing}  & 25.6 & \underline{25.4} & 50.4 & 52.6 & 80.3 & \underline{82.4} & 35.6 & \textbf{39.1} \\
    \rowcolor{gray!20}
    \quad + \textit{d-TreeRPO (ours)}  & \textbf{92.9} \textcolor{darkgreen}{\huge(+86.2)} & \textbf{80.3} \textcolor{darkgreen}{\huge(+74.8)} & \textbf{71.1} \textcolor{darkgreen}{\huge(+51.6)} & \textbf{62.1} \textcolor{darkgreen}{\huge(+46.1)} & \textbf{81.2} \textcolor{darkgreen}{\huge(+4.5)} & \textbf{82.6} \textcolor{darkgreen}{\huge(+3.6)} & \textbf{37.7} \textcolor{darkgreen}{\huge(+5.3)} & 38.9 \textcolor{darkgreen}{\huge(+2.7)} \\
    
    \midrule 
    LLaDA-MoE-7BA1B-Instruct & 12.7 \textcolor{gray}{\huge(+0.0)} & 8.1 \textcolor{gray}{\huge(+0.0)} & 42.6 \textcolor{gray}{\huge(+0.0)} & 41.4 \textcolor{gray}{\huge(+0.0)} & 71.8 \textcolor{gray}{\huge(+0.0)} & 71.0 \textcolor{gray}{\huge(+0.0)} & 30.1 \textcolor{gray}{\huge(+0.0)} & 42.2 \textcolor{gray}{\huge(+0.0)}\\
    \quad \textit{+ Diffu-GRPO}~\citep{zhao2025d1}  & 17.3 & 15.4 & 50.1 & 48.2 & 74.2 & 73.7 & 38.1 & 43.4 \\
    \quad + \textit{wd1}~\citep{tang2025wd1}  & \underline{40.6} & \underline{35.4} & 56.6 & \underline{58.7} & \textbf{76.2} & 76.4 & 39.8 & 43.4 \\
    \quad + \textit{SAPO}~\citep{xie2025step}  & 21.2 & 18.7 & 54.2 & 54.8 & 75.2 & \underline{76.8} & 38.6 & 40.4 \\
    \quad + \textit{GDPO}~\citep{rojas2025improving} & 31.2 & 30.6 & \underline{58.1} & 53.4 & 74.8 & 75.2 & 38.4 & 41.2 \\
    \quad + \textit{TraceRL}~\citep{wang2025revolutionizing}  & 27.4 &  25.2 & 54.2 & 49.1 & 75.4 & 76.1 & \underline{40.0} & \underline{44.1} \\
    \rowcolor{gray!20}
    \quad + \textit{d-TreeRPO (ours)}  & \textbf{78.3} \textcolor{darkgreen}{\huge(+65.6)} & \textbf{72.0} \textcolor{darkgreen}{\huge(+63.9)} & \textbf{67.2} \textcolor{darkgreen}{\huge(+24.6)} & \textbf{60.6} \textcolor{darkgreen}{\huge(+19.2)} & \underline{75.5} \textcolor{darkgreen}{\huge(+3.7)} & \textbf{77.4} \textcolor{darkgreen}{\huge(+6.4)} & \textbf{41.2} \textcolor{darkgreen}{\huge(+11.1)} & \textbf{46.3} \textcolor{darkgreen}{\huge(+4.1)} \\
    
    \bottomrule
  \end{tabular}
  } 
  \vspace{-5pt}
  \caption{Performance comparison of \emph{d}-TreeRPO with existing dLLM RL methods on reasoning tasks under 256/512-token generation settings. \textbf{Bold} values indicate the best results, with the second-best \underline{underlined}. Performance gains of \emph{d}-TreeRPO relative to the base model are shown in parentheses.}
  \vspace{-5pt}
  \label{tab:main_results}
\end{table*}

\subsection{Experimental Setup}
\textbf{Models and Datasets.} We employ LLaDA-8B-Instruct~\citep{nie2025llada} and LLaDA-MoE~\citep{zhu2025llada} as the base model\footnote{We did not use Dream~\citep{ye2025dream} because our experiments found its output format inconsistent, which caused instability across various dLLM RL algorithms.}. For training and evaluation, we utilize two puzzle-based reasoning tasks: Sudoku and Countdown, and two mathematical reasoning tasks: GSM8K~\citep{cobbe2021gsm8k} and Math500~\citep{lightman2023lets}.

\vspace{2pt}

\noindent\textbf{Baselines.} Evaluated dLLM RL baselines include Diffu-GRPO~\citep{zhao2025d1}, VRPO (LLaDA-1.5)~\citep{zhu2025llada1_5}, wd1~\citep{tang2025wd1}, SAPO~\citep{xie2025step}, GDPO~\citep{rojas2025improving} and TraceRL~\citep{wang2025revolutionizing}. Detailed descriptions of each algorithms are provided in Appendix \ref{app:baselines}.

\vspace{2pt}

\noindent\textbf{Training Details.} Following previous works~\citep{zhao2025d1,tang2025wd1}, we use LoRA training with rank $r=128$ and scaling factor $\alpha=64$. We adopt a learning rate of $3 \times 10^{-5}$, set $\tau_{\text{max}}=2$ and $\beta=0.7$ in Eq.~(\ref{eq:tau-t}), and set $\lambda_{\text{max}}=0.003$ and $\gamma=2$ in Eq.~(\ref{eq:lambda-1}) (see Appendix \ref{app:hyper} for hyper-parameter analysis). The maximum generation length is set to 256 tokens with block-wise decoding (block length$=32$) over 128 denoising steps. We set temperature$=0.9$ and use confidence-based decoding during rollout. To balance computational efficiency and performance, the tree-structured rollout employs depth $H=2$ and branch factor $B=4$ (see Section~\ref{sec:hyper} for sensitivity analysis). More training details are provided in Appendix~\ref{app:training-details}.


\noindent\textbf{Evaluation Details.} We evaluate model performance under 256-token and 512-token outputs, with block-wise decoding (block length$=32$) and temperature=$0.0$. Denoising steps are configured as half the generation length. All tasks adopt zero-shot evaluation with pass@1 scoring.

\vspace{-4pt}
\subsection{Main Results}
\label{sec:exp-main}
\vspace{-2pt}
\textbf{Overall Performance}. Table~\ref{tab:main_results} presents a comparison between \emph{d}-TreeRPO and existing dLLM RL methods, using LLaDA-8B-Instruct and LLaDA-MoE-7BA1B-Instruct as the base models. Our approach achieves state-of-the-art performance in most settings, showcasing substantial improvements over the base models: +\textbf{86.2\%} on Sudoku, +\textbf{51.6\%} on Countdown, +\textbf{4.5\%} on GSM8K, and +\textbf{5.3\%} on Math500 with 256-token generations on LLaDA-8B-Instruct, as well as +\textbf{65.6\%} on Sudoku, +\textbf{24.6\%} on Countdown, +\textbf{3.7\%} on GSM8K, and +\textbf{11.1\%} on Math500 with 256-token generations on LLaDA-MoE-7BA1B-Instruct. Appendix \ref{app:dynamics-1} demonstrates the training reward curves. Furthermore, case studies are provided in Appendix \ref{app:case}.

\begin{figure}[t]
    \centering
    \includegraphics[width=\linewidth]{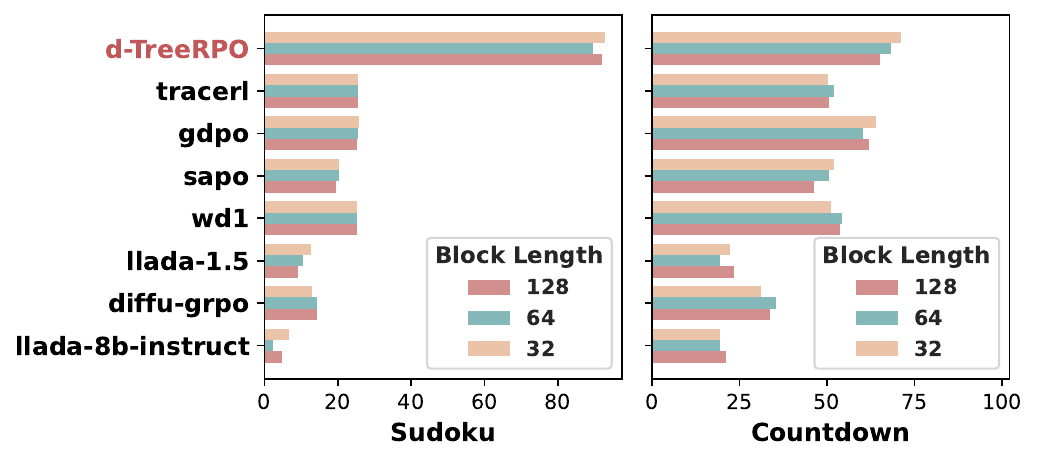}
    \vspace{-20pt}
    \caption{Performance comparison of \emph{d}-TreeRPO with dLLM RL baselines under different decoding strategies.}
    \label{fig:robustness}
    \vspace{-20pt}
\end{figure}

\noindent\textbf{Robustness Against Decoding Strategies.}
Figure \ref{fig:robustness} shows the performance of \emph{d}-TreeRPO and dLLM RL baselines under different block-wise decoding strategies. Using LLaDA-8B-Instruct as the base model with 256-token generations, experiments on Sudoku and Countdown tasks were conducted with block lengths of 32, 64, and 128 for both training and inference. \emph{d}-TreeRPO consistently outperforms baselines across all strategies, demonstrating strong robustness.

\begin{figure}[t] 
    \centering
    \begin{subfigure}[t]{0.49\linewidth} 
        \centering
        \includegraphics[width=\linewidth]{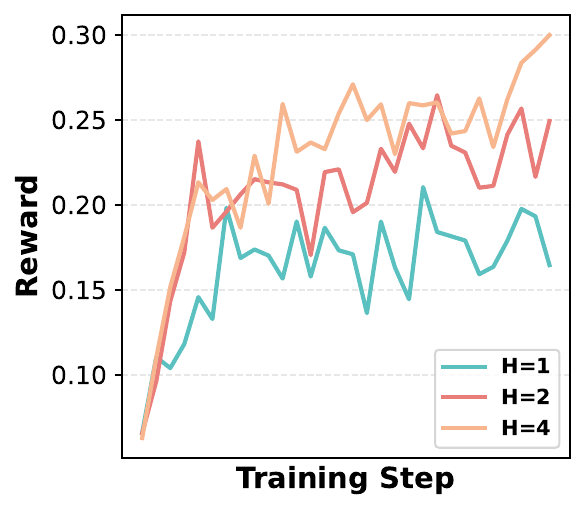}
        \caption{$H$}
        \label{fig:H}
    \end{subfigure}
    \hfill 
    \begin{subfigure}[t]{0.49\linewidth}
        \centering
        \includegraphics[width=\linewidth]{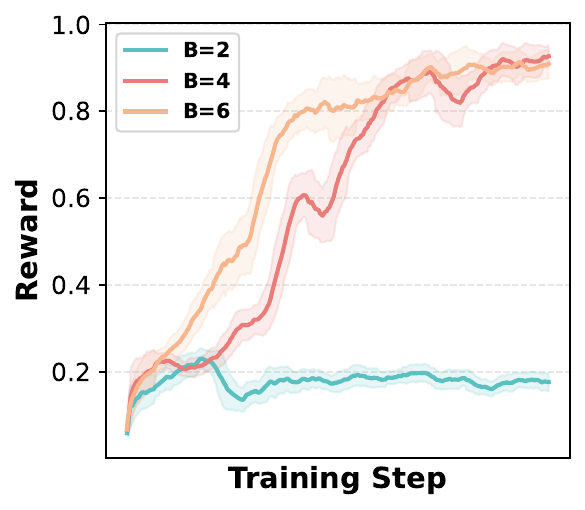}
        \caption{$B$}
        \label{fig:B}
    \end{subfigure}
    \caption{Training rewards on Sudoku task under different parameters $H$ and $B$ (LLaDA-8B-Instruct).}
    \label{fig:combined}
\end{figure}

\vspace{-2pt}

\subsection{Comparing \emph{d}-TreeRPO with Sparse Outcome Reward Variants under the Same Rollout Budget}
To verify that the gains from d-TreeRPO’s tree-based advantage computation come from finer-grained credit assignment, rather than from using a larger rollout budget, we conduct comparisons under the same rollout budget, i.e., the same number of leaf completions $B^H=4^2=16$. Specifically, we compare:
\begin{itemize}
    \item \textbf{Ours:} tree-structured rollout with bottom-up value estimation and parent-to-child (step-wise) fine-grained advantages.
    \item \textbf{Sparse-outcome-reward variants:} directly broadcasting the final outcome reward to all tokens, evaluated under two rollout protocols with the identical budget: \textbf{(i) Sparse-Tree:} the same tree-structured rollout ($B^H=4^2=16$ leaf nodes) but without bottom-up credit assignment, and \textbf{(ii) Sparse-Flat:} independently sampling $B^H=4^2=16$ full completions (no tree).
\end{itemize}

\begin{table}[t]
    \centering
    \resizebox{\linewidth}{!}{
    \begin{tabular}{lcccc}
    \toprule
        \textbf{Method} & \textbf{Sudoku} & \textbf{Countdown} & \textbf{GSM8k} & \textbf{Math500} \\
        \midrule
        \rowcolor{gray!30}
        \emph{d}-TreeRPO & \textbf{92.9} & \textbf{71.1} & \textbf{81.2} & \textbf{37.7} \\
        Sparse-Tree & 22.4 & 38.2 & 80.4 & 35.4 \\
        Sparse-Flat & 24.6 & 37.1 & 80.4 & 35.6 \\
        \bottomrule
    \end{tabular}
    }
    \vspace{-5pt}
    \caption{Comparison of d-TreeRPO with two sparse-outcome-reward variants (Sparse-Tree and Sparse-Flat) on four tasks under the same rollout budget ($B^H=4^2=16$ per query), using LLaDA-8B-Instruct with 256-token generations.}
    \label{tab:sparse}
    \vspace{-10pt}
\end{table}

\noindent The results are shown in Table \ref{tab:sparse}. As can be seen, under an identical rollout budget, the sparse-outcome-reward variants (Sparse-Tree and Sparse-Flat) perform consistently worse than d-TreeRPO. The gap is especially large on the puzzle-based tasks, where sparse reward broadcasting falls far behind, highlighting the importance of fine-grained credit assignment.

\subsection{Tree Height and Branch Factor Analysis}
\label{sec:hyper}
\vspace{-2pt}
\textbf{Tree Height $H$ Analysis.} Tree height $H$ represents the steps required to traverse from the root to a leaf node. Increasing $H$ enables finer granularity in generation, improving reward precision. Training curves for $H=1, 2, 4$ over the first 30 batches (Figure \ref{fig:H}) show that $H=4$ improves the fastest, followed by $H=2$, while $H=1$ progresses the slowest, validating our analysis. However, the exponential computational cost of \emph{d}-TreeRPO at higher $H$ prevented $H=4$ from converging within reasonable time given our resource constraints (see Section \ref{sec:compute} for details). Therefore, we adopted $H=2$ as the primary setting for our experiments, balancing efficiency and performance.

\vspace{2pt}

\noindent\textbf{Branch Factor $B$ Analysis.}  
The branch factor $B$ represents the number of child nodes expanded from a parent during rollout. A larger $B$ improves exploration diversity and reduces parent node reward variance. Training curves for $B=2$, $4$, and $6$ (Figure~\ref{fig:B}) show that $B=2$ achieves a final reward of only $0.2$, while $B=4$ and $B=6$ converge above $0.9$, with $B=6$ showing faster initial improvement. However, due to polynomial growth in computational cost with $B$ (Section~\ref{sec:compute}), we use $B=4$ in our main experiments.

\vspace{-5pt}
\subsection{Effectiveness of Self-distillation Loss}
\vspace{-2pt}
\label{sec:exp-effective} 
\begin{figure}[t]
    \centering
    \begin{subfigure}[t]{0.49\linewidth}
        \centering
        \includegraphics[width=\textwidth]{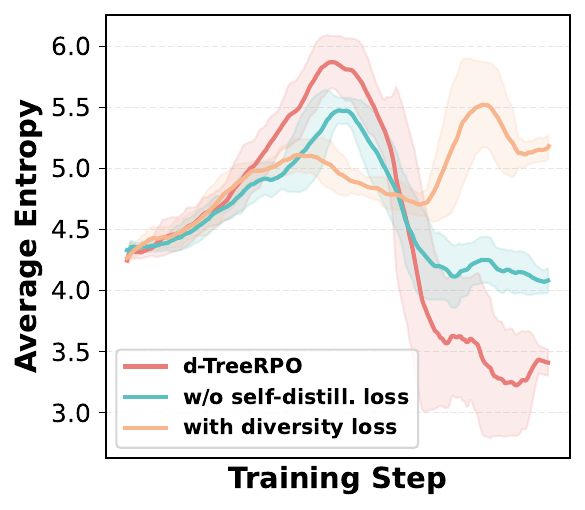} 
        \caption{Entropy curves}
        \label{fig:sudoku-entropy}
    \end{subfigure}
    \hfill 
    \begin{subfigure}[t]{0.49\linewidth}
        \centering
        \includegraphics[width=\textwidth]{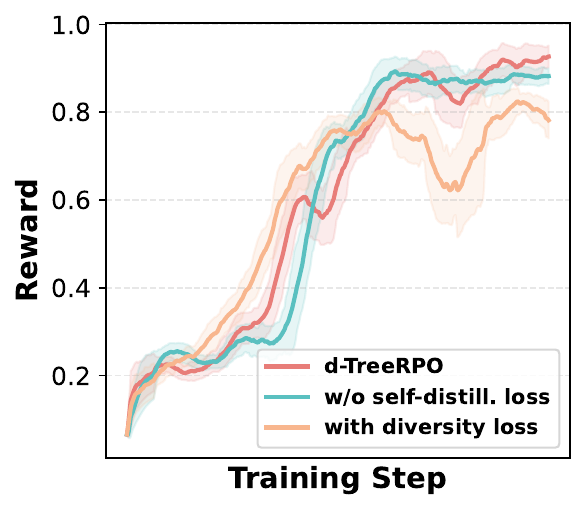} 
        \caption{Reward curves}
        \label{fig:sudoku-reward}
    \end{subfigure}
    \caption{Training curves for the Sudoku task under different ablations of self-distillation loss (LLaDA-8B-Instruct).}
    \label{fig:sudoku_curves}
    \vspace{-8pt}
\end{figure}

\begin{table}[t]
    \centering
    \resizebox{\linewidth}{!}{
    \begin{tabular}{lcccc}
        \toprule
        \textbf{Method} & \textbf{Sudoku} & \textbf{Countdown} & \textbf{GSM8k} & \textbf{Math500} \\
        \midrule
        \rowcolor{gray!30}
        \emph{d}-TreeRPO & \textbf{92.9} & \textbf{71.1} & \textbf{81.2} & \textbf{37.7} \\
        \emph{d}-TreeRPO w/o distill. & 89.8 & 66.4 & 80.9 & 36.1 \\
        \emph{d}-TreeRPO w. div. & 84.2 & 63.4 & 78.5 & 35.2\\
        \bottomrule
    \end{tabular}
    }
    \label{tab:final_performance}
    \vspace{-5pt}
    \caption{Comparison of evaluation performance of \emph{d}-TreeRPO, \emph{d}-TreeRPO without self-distillation loss and \emph{d}-TreeRPO with diversity-promoting loss, using LLaDA-8B-Instruct with 256-token generations.}
    \label{tab:final_performance}
\end{table}

\textbf{Setup}. To validate the effectiveness of the proposed self-distillation loss, we perform ablation studies. First, we compare the full \emph{d}-TreeRPO model to a variant without the self-distillation loss. Additionally, to highlight the importance of progressively sharpening policy distribution, we introduce the \textbf{diversity-promoting loss} ($\mathcal{L}_{\text{div}}$). Designed as a symmetric counterpart to $\mathcal{L}_{\text{distill}}$, $\mathcal{L}_{\text{div}}$ intentionally increases policy diversity by reducing model confidence in later training stages. Inspired by prior work~\citep{yang2025entropic}, the loss increases entropy for negative samples. It operates on child nodes with negative advantages ($C_p^- = \{c \in C_p \mid A_p^c < 0\}$), assigning weights proportional to $|A_p^c|$. A negative target distribution, $P_{\text{div}}$, is constructed through weighted voting. Rather than minimizing KL divergence, the policy is trained to move away from this target via a negative KL objective, using the same time-scheduling coefficient $\lambda(t)$ as $\mathcal{L}_{\text{distill}}$. Therefore, $\mathcal{L}_\text{diversity}$ is:
\begin{equation}
\scriptsize
\label{eq:L-div}
- \lambda(t) \cdot \mathbb{E}_{q \sim \mathcal{D},\;p \in \mathcal{T}(q)} \left[ \frac{1}{k}\sum_{i=1}^k D_{\text{KL}} \left( P_{\text{div}}^{\sigma_i}(\cdot) \parallel \pi_\theta^{\sigma_i}(\cdot|p) \right) \right].
\end{equation}

\begin{table}[t]
\renewcommand{\arraystretch}{1.0}
\Huge
    \centering
    \label{tab:estimation-error}
    \resizebox{\linewidth}{!}{
    \begin{tabular}{lcccc}
        \toprule
        \textbf{Method} & \textbf{Sudoku} & \textbf{Countdown} & \textbf{GSM8k} & \textbf{Math500} \\
        \midrule
        \rowcolor{gray!30}
         \emph{d}-TreeRPO & 1.25 $\pm$ {\huge 1.15}  &  1.45 $\pm$ {\huge 1.21} & 1.32 $\pm$ {\huge 0.97} & 1.42 $\pm$ {\huge 1.14}\\
         \emph{d}-TreeRPO w/o distill. & 2.64 $\pm$ {\huge 1.97} & 1.61 $\pm$ {\huge 1.30} & 1.40 $\pm$ {\huge 1.01} & 1.76 $\pm$ {\huge 1.32}\\
         \emph{d}-TreeRPO w. div & 2.83 $\pm$ {\huge 2.02}& 1.66 $\pm$ {\huge 1.24} & 1.43 $\pm$ {\huge 1.01} & 1.82 $\pm$ {\huge 1.26}\\
        \bottomrule
    \end{tabular}
    }
    \vspace{-10pt}
    \caption{Comparison of probability estimation error $\log (p_{\text{true}}/\hat{p})$ of \emph{d}-TreeRPO, \emph{d}-TreeRPO without self-distillation loss and \emph{d}-TreeRPO with diversity-promoting loss (LLaDA-8B-Instruct).}
    \label{tab:estimate-error}
    \vspace{-5pt}
\end{table}

\begin{figure*}[t]
    \centering 

    \begin{subfigure}[t]{0.242\textwidth}
        \centering
        \includegraphics[width=\textwidth]{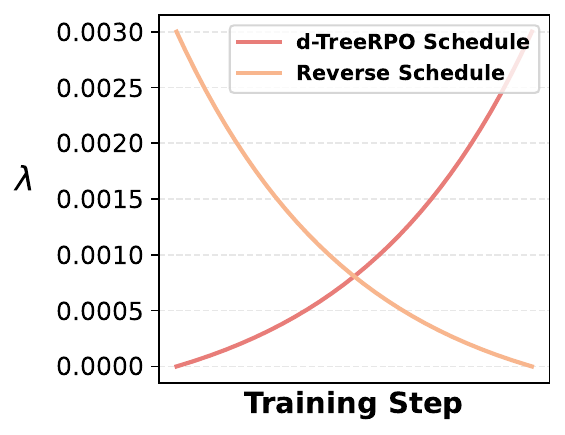} 
        \caption{$\lambda$}
        \label{fig:lambda}
    \end{subfigure}
    \hfill 
    \begin{subfigure}[t]{0.24\textwidth}
        \centering
        \includegraphics[width=\textwidth]{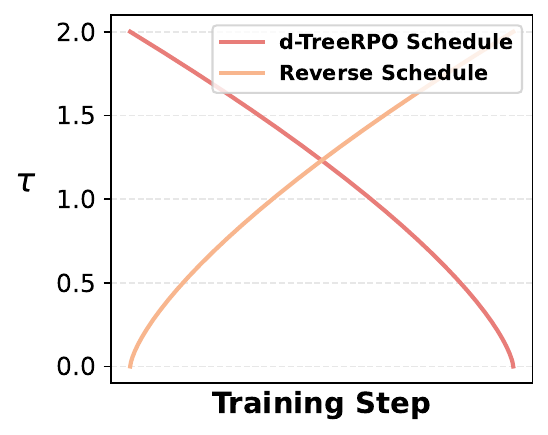} 
        \caption{$\tau$}
        \label{fig:tau}
    \end{subfigure}
    \hfill 
    \begin{subfigure}[t]{0.254\textwidth}
        \centering
        \includegraphics[width=\textwidth]{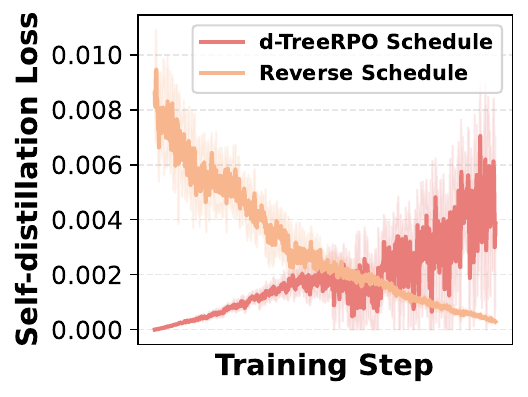} 
        \caption{Self-distillation loss}
        \label{fig:self-distill-loss}
    \end{subfigure}
    \begin{subfigure}[t]{0.246\textwidth}
        \centering
        \includegraphics[width=\textwidth]{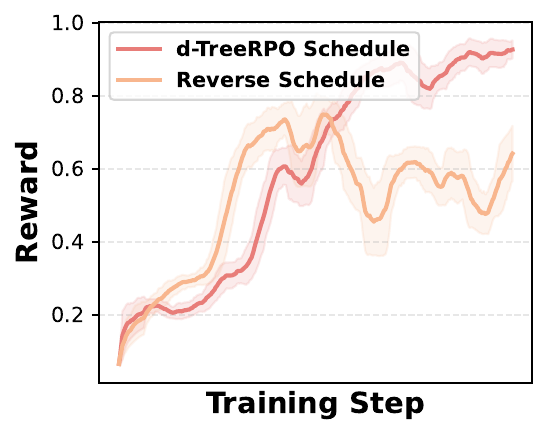} 
        \caption{Reward}
        \label{fig:reward}
    \end{subfigure}
    \vspace{-4pt}
    \caption{Training dynamics comparison of \emph{d}-TreeRPO and its reverse-scheduled variant on the Sudoku task with LLaDA-8B-Instruct as the base model.}
    \label{fig:reverse}
    \vspace{-5pt}
\end{figure*}

\noindent\textbf{Self-distillation Loss Enhance Determinism}. Figure~\ref{fig:sudoku_curves} shows policy entropy and reward trends for the above three settings on the Sudoku task (additional results on more task is in Appendix \ref{app:dynamics-2}). Initially, all models exhibit increasing entropy, typical of the exploration phase. Later, model with self-distillation loss sharply reduces entropy, improving policy confidence. In contrast, model with diversity-promoting loss maintains high entropy, reflecting increased variability. Correspondingly, model trained with the complete \emph{d}-TreeRPO achieves the highest reward, followed by the the model without self-distillation loss, while the diversity-focused variant has the lowest reward and unstable training behavior.

\noindent\textbf{Self-distillation Loss Reduces Estimation Error.}
We estimate Eq.~(\ref{eq:unbiased}) via Monte Carlo by sampling $32$ random decoding orders per sample and computing, for each token, the probability of the realized token at the step when it is revealed; this yields $p_{\mathrm{true}}$ and the per-token error $\log\!\big(p_{\mathrm{true}}/\hat{p}\big)$.
We evaluate converged checkpoints trained with and without self-distillation, as well as with diversity-promoting loss on four tasks, using $100$ samples per task, and report the mean and standard deviation of $\log\!\big(p_{\mathrm{true}}/\hat{p}\big)$. As shown in Table \ref{tab:estimate-error}, \emph{d}-TreeRPO with self-distillation loss achieve the smallest estimation error, proving its effectiveness.

\noindent\textbf{Performance Results.} Furthermore, we report the final evaluation performance of these three settings across all four tasks in Table~\ref{tab:final_performance}. 
The results align with our training-time observations: the full \emph{d}-TreeRPO algorithm consistently outperforms the other two variants. Removing the self-distillation loss leads to a noticeable drop in performance, while adding the diversity-promoting loss has the worst performance. These findings provide compelling evidence for the practical effectiveness of our proposed self-distillation loss. 

\vspace{-4pt}
\subsection{Analysis of Time-scheduled Design}
\vspace{-3pt}
\label{sec:time-scheduled}
\textbf{Comparison with Reverse-Scheduled Self-Distillation Loss.}  
To assess the design choice of increasing $\lambda(t)$ and decreasing $\tau(t)$ in our self-distillation loss, we tested a reverse schedule for $\lambda(t)$ and $\tau(t)$. Figure~\ref{fig:reverse} compares the forward schedule used in \emph{d}-TreeRPO with the reverse schedule, showing the evolution of $\lambda$, $\tau$, self-distillation loss, and reward during training. While the reverse schedule initially yields faster reward growth due to a strong distillation effect, its performance drops after reaching a reward of 0.75, highlighting insufficient generalization. This analysis supports the forward scheduling used in \emph{d}-TreeRPO. 

\vspace{-6pt}
\begin{table}[t]
\vspace{-5pt}
\renewcommand{\arraystretch}{0.9}
\huge
\centering
\resizebox{\linewidth}{!}{
\begin{tabular}{lcccc}
\toprule
\textbf{Method} & \textbf{Batch (s)} & \textbf{Update (s)} & \textbf{Conv. (h)} & \textbf{Acc} (\%) \\
\midrule
Diffu-GRPO & 109 & 9.08 & $\approx$24 & 12.9 \\
wd1        & 87  & 7.25 & $\approx$24 & 25.2 \\
SAPO       & 423 & 35.25& $\approx$72 & 20.3 \\
GDPO       & 153 & 12.72 & $\approx$48 & 25.7 \\
TraceRL    & 604 & 43.14 & $\approx$48 & 25.6 \\
\cellcolor{gray!30}\emph{d}-TreeRPO & \cellcolor{gray!30}598 & \cellcolor{gray!30}9.96 & \cellcolor{gray!30}$\approx$48 & \cellcolor{gray!30}92.9 \\
\bottomrule
\end{tabular}
}
\vspace{-5pt}
\caption{Training time comparison between \emph{d}-TreeRPO and other dLLM RL baselines on the Sudoku task using LLaDA-8B-Instruct as base model, including the training time for processing a batch, the time per parameter update, and total time to convergence. All methods are tested under 8 H20 GPUs, batch size = 4/device.}
\vspace{-5pt}
\label{tab:time-cost}
\end{table}

\subsection{Computational Overhead}
\vspace{-2pt}
\label{sec:compute}
\textbf{Computational Analysis of \emph{d}-TreeRPO.}  
The computational cost of \emph{d}-TreeRPO stems mainly from two components: (1)~the rollout phase and (2)~loss computation with gradient updates. For the rollout phase, a tree with height $H$ and branch factor $B$ involves $\sum_{i=1}^{H} B^i = \frac{B(B^H - 1)}{B - 1}$ tree steps. With $N$ denoising steps for a complete generation, each tree step includes $N/H$ denoising steps. Thus, constructing a single tree requires $\left( \frac{B(B^H - 1)}{B - 1} \right) \cdot \frac{N}{H}$ denoising steps. During loss computation and gradient updates, $\frac{B(B^H - 1)}{B - 1}$ forward passes are needed. To optimize memory usage, we compute loss for one subtree at a time during each update.

\vspace{2pt}

\noindent\textbf{Training Time Comparison.}  
To demonstrate the efficiency of \emph{d}-TreeRPO, we compared its training time with other dLLM RL baselines under the same GPU setup (8 H20 GPUs) and train batch size (4 per device). As shown in Table~\ref{tab:time-cost}, \emph{d}-TreeRPO achieves moderate training speed: slower than Diffu-GRPO and wd1, comparable to TraceRL and GDPO, but faster than SAPO. Despite this, its final performance surpasses all baselines significantly, making it a practical and effective choice.

\vspace{-5pt}
\section{Conclusion}
\vspace{-4pt}
We introduced \emph{d}-TreeRPO, a reliable RL algorithm for dLLMs. \emph{d}-TreeRPO organizes rollouts as trees and propagates rewards bottom-up to acquire granular advantage signals. The incorporation of a self-distillation loss enhances determinism in later training stages and reduces estimation error. Experimental results demonstrate that \emph{d}-TreeRPO consistently outperforms baselines across multiple reasoning tasks, proving its effectiveness and practicality.

\section*{Limitations}

While \emph{d}-TreeRPO provides a reliable RL framework for dLLMs, several limitations remain. Our experiments primarily focus on reasoning tasks with automatically verifiable outcome rewards (e.g., exact-solution puzzles and correctness-graded math problems), where \emph{d}-TreeRPO can propagate reliable signals through tree-structured rollouts. Extending the framework to settings with less verifiable or more subjective objectives and studying how to construct reliable intermediate credit assignment in such cases remain important future directions. In addition, our evaluation is conducted in a zero-shot manner and mainly reports pass@1 score. Incorporating broader evaluation protocols would provide a more comprehensive assessment.

\section*{Acknowledgments} 
This work is primarily supported by the Key Research and Development Program of China (No. 2024YFB3309702). We would like to express our gratitude to the anonymous ARR January reviewers (Reviewer EPXK, Q6fg, tKJg, UkBS) and Area Chairs (Area Chair qvMv and ftyK) for their valuable feedback and suggestions that helped improve this paper. 


\bibliography{custom}

\begin{thebibliography}{38}
\providecommand{\natexlab}[1]{#1}

\bibitem[{Arriola et~al.(2025)Arriola, Gokaslan, Chiu, Yang, Qi, Han, Sahoo, and Kuleshov}]{arriola2025block}
Marianne Arriola, Aaron Gokaslan, Justin~T Chiu, Zhihan Yang, Zhixuan Qi, Jiaqi Han, Subham~Sekhar Sahoo, and Volodymyr Kuleshov. 2025.
\newblock Block diffusion: Interpolating between autoregressive and diffusion language models.
\newblock \emph{arXiv preprint arXiv:2503.09573}.

\bibitem[{Cobbe et~al.(2021)Cobbe, Kosaraju, Bavarian, Chen, Jun, Kaiser, Plappert, Tworek, Hilton, Nakano, Hesse, and Schulman}]{cobbe2021gsm8k}
Karl Cobbe, Vineet Kosaraju, Mohammad Bavarian, Mark Chen, Heewoo Jun, Lukasz Kaiser, Matthias Plappert, Jerry Tworek, Jacob Hilton, Reiichiro Nakano, Christopher Hesse, and John Schulman. 2021.
\newblock Training verifiers to solve math word problems.
\newblock \emph{arXiv preprint arXiv:2110.14168}.

\bibitem[{Cui et~al.(2025)Cui, Zhang, Chen, Yuan, Wang, Zuo, Li, Fan, Chen, Chen et~al.}]{cui2025entropy}
Ganqu Cui, Yuchen Zhang, Jiacheng Chen, Lifan Yuan, Zhi Wang, Yuxin Zuo, Haozhan Li, Yuchen Fan, Huayu Chen, Weize Chen, and 1 others. 2025.
\newblock The entropy mechanism of reinforcement learning for reasoning language models.
\newblock \emph{arXiv preprint arXiv:2505.22617}.

\bibitem[{Fang et~al.(2026)Fang, Liu, Zou, Chen, Ma, Pan, Miao, Huang, Liu, and Yu}]{fang2026locally}
Liancheng Fang, Aiwei Liu, Henry~Peng Zou, Yankai Chen, Enze Ma, Leyi Pan, Chunyu Miao, Wei-Chieh Huang, Xue Liu, and Philip~S Yu. 2026.
\newblock Locally confident, globally stuck: The quality-exploration dilemma in diffusion language models.
\newblock \emph{arXiv preprint arXiv:2604.00375}.

\bibitem[{Gong et~al.(2025)Gong, Zhang, Zheng, Gu, Jaitly, Kong, and Zhang}]{gong2025diffucoder}
Shansan Gong, Ruixiang Zhang, Huangjie Zheng, Jiatao Gu, Navdeep Jaitly, Lingpeng Kong, and Yizhe Zhang. 2025.
\newblock Diffucoder: Understanding and improving masked diffusion models for code generation.
\newblock \emph{arXiv preprint arXiv:2506.20639}.

\bibitem[{Huang et~al.(2025)Huang, Chen, Wang, Li, and Qi}]{huang2025reinforcing}
Zemin Huang, Zhiyang Chen, Zijun Wang, Tiancheng Li, and Guo-Jun Qi. 2025.
\newblock Reinforcing the diffusion chain of lateral thought with diffusion language models.
\newblock \emph{arXiv preprint arXiv:2505.10446}.

\bibitem[{Li et~al.(2025)Li, Callanan, Zhu, Sibue, Papadimitriou, Mahfouz, Ma, and Liu}]{li2025entropy}
Xianzhi Li, Ethan Callanan, Xiaodan Zhu, Mathieu Sibue, Antony Papadimitriou, Mahmoud Mahfouz, Zhiqiang Ma, and Xiaomo Liu. 2025.
\newblock Entropy-aware branching for improved mathematical reasoning.
\newblock \emph{arXiv preprint arXiv:2503.21961}.

\bibitem[{Lightman et~al.(2023)Lightman, Kosaraju, Burda, Edwards, Baker, Lee, Leike, Schulman, Sutskever, and Cobbe}]{lightman2023lets}
Hunter Lightman, Vineet Kosaraju, Yura Burda, Harri Edwards, Bowen Baker, Teddy Lee, Jan Leike, John Schulman, Ilya Sutskever, and Karl Cobbe. 2023.
\newblock Let's verify step by step.
\newblock \emph{arXiv preprint arXiv:2305.20050}.

\bibitem[{Lin et~al.(2025)Lin, Zhang, Hou, and Li}]{lin2025boundary}
Nianyi Lin, Jiajie Zhang, Lei Hou, and Juanzi Li. 2025.
\newblock Boundary-guided policy optimization for memory-efficient rl of diffusion large language models.
\newblock \emph{arXiv preprint arXiv:2510.11683}.

\bibitem[{Liu et~al.(2025)Liu, He, Zeng, Zhang, Zhang, Wu, Jia, Liu, Zhou, and Zhou}]{liu2025wedlm}
Aiwei Liu, Minghua He, Shaoxun Zeng, Sijun Zhang, Linhao Zhang, Chuhan Wu, Wei Jia, Yuan Liu, Xiao Zhou, and Jie Zhou. 2025.
\newblock Wedlm: Reconciling diffusion language models with standard causal attention for fast inference.
\newblock \emph{arXiv preprint arXiv:2512.22737}.

\bibitem[{Nie et~al.(2024)Nie, Zhu, Du, Pang, Liu, Zeng, Lin, and Li}]{nie2024scaling}
Shen Nie, Fengqi Zhu, Chao Du, Tianyu Pang, Qian Liu, Guangtao Zeng, Min Lin, and Chongxuan Li. 2024.
\newblock Scaling up masked diffusion models on text.
\newblock \emph{arXiv preprint arXiv:2410.18514}.

\bibitem[{Nie et~al.(2025)Nie, Zhu, You, Zhang, Ou, Hu, Zhou, Lin, Wen, and Li}]{nie2025llada}
Shen Nie, Fengqi Zhu, Zebin You, Xiaolu Zhang, Jingyang Ou, Jun Hu, Jun Zhou, Yankai Lin, Ji-Rong Wen, and Chongxuan Li. 2025.
\newblock Large language diffusion models.
\newblock \emph{arXiv preprint arXiv:2502.09992}.

\bibitem[{Ou et~al.(2025)Ou, Han, Xu, Xu, Xie, Ermon, Wu, and Li}]{ou2025principled}
Jingyang Ou, Jiaqi Han, Minkai Xu, Shaoxuan Xu, Jianwen Xie, Stefano Ermon, Yi~Wu, and Chongxuan Li. 2025.
\newblock Principled rl for diffusion llms emerges from a sequence-level perspective.
\newblock \emph{arXiv preprint arXiv:2512.03759}.

\bibitem[{Ou et~al.(2024)Ou, Nie, Xue, Zhu, Sun, Li, and Li}]{ou2024your}
Jingyang Ou, Shen Nie, Kaiwen Xue, Fengqi Zhu, Jiacheng Sun, Zhenguo Li, and Chongxuan Li. 2024.
\newblock Your absorbing discrete diffusion secretly models the conditional distributions of clean data.
\newblock \emph{arXiv preprint arXiv:2406.03736}.

\bibitem[{Rojas et~al.(2025)Rojas, Lin, Rasul, Schneider, Nevmyvaka, Tao, and Deng}]{rojas2025improving}
Kevin Rojas, Jiahe Lin, Kashif Rasul, Anderson Schneider, Yuriy Nevmyvaka, Molei Tao, and Wei Deng. 2025.
\newblock Improving reasoning for diffusion language models via group diffusion policy optimization.
\newblock \emph{arXiv preprint arXiv:2510.08554}.

\bibitem[{Schulman et~al.(2017)Schulman, Wolski, Dhariwal, Radford, and Klimov}]{schulman2017proximal}
John Schulman, Filip Wolski, Prafulla Dhariwal, Alec Radford, and Oleg Klimov. 2017.
\newblock Proximal policy optimization algorithms.
\newblock \emph{arXiv preprint arXiv:1707.06347}.

\bibitem[{Shao et~al.(2024)Shao, Wang, Zhu, Xu, Song, Bi, Zhang, Zhang, Li, Wu et~al.}]{shao2024deepseekmath}
Zhihong Shao, Peiyi Wang, Qihao Zhu, Runxin Xu, Junxiao Song, Xiao Bi, Haowei Zhang, Mingchuan Zhang, YK~Li, Yang Wu, and 1 others. 2024.
\newblock Deepseekmath: Pushing the limits of mathematical reasoning in open language models.
\newblock \emph{arXiv preprint arXiv:2402.03300}.

\bibitem[{Song et~al.(2025)Song, Zhang, Luo, Gao, Xia, Luo, Li, Yang, Yu, Qu et~al.}]{song2025seed}
Yuxuan Song, Zheng Zhang, Cheng Luo, Pengyang Gao, Fan Xia, Hao Luo, Zheng Li, Yuehang Yang, Hongli Yu, Xingwei Qu, and 1 others. 2025.
\newblock Seed diffusion: A large-scale diffusion language model with high-speed inference.
\newblock \emph{arXiv preprint arXiv:2508.02193}.

\bibitem[{Tan et~al.(2025)Tan, Pan, Lin, Chen, Zheng, Tang, and Yang}]{tan2025gtpo}
Hongze Tan, Jianfei Pan, Jinghao Lin, Tao Chen, Zhihang Zheng, Zhihao Tang, and Haihua Yang. 2025.
\newblock Gtpo and grpo-s: Token and sequence-level reward shaping with policy entropy.
\newblock \emph{arXiv preprint arXiv:2508.04349}.

\bibitem[{Tang et~al.(2025)Tang, Dolga, Yoon, and Bogunovic}]{tang2025wd1}
Xiaohang Tang, Rares Dolga, Sangwoong Yoon, and Ilija Bogunovic. 2025.
\newblock wd1: Weighted policy optimization for reasoning in diffusion language models.
\newblock \emph{arXiv preprint arXiv:2507.08838}.

\bibitem[{Wang et~al.(2025{\natexlab{a}})Wang, Schiff, Turok, and Kuleshov}]{wang2025d2}
Guanghan Wang, Yair Schiff, Gilad Turok, and Volodymyr Kuleshov. 2025{\natexlab{a}}.
\newblock d2: Improved techniques for training reasoning diffusion language models.
\newblock \emph{arXiv preprint arXiv:2509.21474}.

\bibitem[{Wang et~al.(2025{\natexlab{b}})Wang, Yang, Li, Tian, Shen, and Wang}]{wang2025revolutionizing}
Yinjie Wang, Ling Yang, Bowen Li, Ye~Tian, Ke~Shen, and Mengdi Wang. 2025{\natexlab{b}}.
\newblock Revolutionizing reinforcement learning framework for diffusion large language models.
\newblock \emph{arXiv preprint arXiv:2509.06949}.

\bibitem[{Xie et~al.(2025{\natexlab{a}})Xie, Kong, Song, Dong, Chen, Xing, and Zhang}]{xie2025step}
Shaoan Xie, Lingjing Kong, Xiangchen Song, Xinshuai Dong, Guangyi Chen, Eric~P Xing, and Kun Zhang. 2025{\natexlab{a}}.
\newblock Step-aware policy optimization for reasoning in diffusion large language models.
\newblock \emph{arXiv preprint arXiv:2510.01544}.

\bibitem[{Xie et~al.(2025{\natexlab{b}})Xie, Ye, Zheng, Gao, Dong, Wu, Zhao, Gong, Jiang, Li et~al.}]{xie2025dreamcoder}
Zhihui Xie, Jiacheng Ye, Lin Zheng, Jiahui Gao, Jingwei Dong, Zirui Wu, Xueliang Zhao, Shansan Gong, Xin Jiang, Zhenguo Li, and 1 others. 2025{\natexlab{b}}.
\newblock Dream-coder 7b: An open diffusion language model for code.
\newblock \emph{arXiv preprint arXiv:2509.01142}.

\bibitem[{Yang et~al.(2025{\natexlab{a}})Yang, Chen, Hu, and Shao}]{yang2025taming}
Jingyi Yang, Guanxu Chen, Xuhao Hu, and Jing Shao. 2025{\natexlab{a}}.
\newblock Taming masked diffusion language models via consistency trajectory reinforcement learning with fewer decoding step.
\newblock \emph{arXiv preprint arXiv:2509.23924}.

\bibitem[{Yang et~al.(2025{\natexlab{b}})Yang, Xu, Chen, Liu, Lyu, Lin, Ye, and Yang}]{yang2025entropic}
Kai Yang, Xin Xu, Yangkun Chen, Weijie Liu, Jiafei Lyu, Zichuan Lin, Deheng Ye, and Saiyong Yang. 2025{\natexlab{b}}.
\newblock Entropic: Towards stable long-term training of llms via entropy stabilization with proportional-integral control.
\newblock \emph{arXiv preprint arXiv:2511.15248}.

\bibitem[{Yang et~al.(2025{\natexlab{c}})Yang, Tian, Li, Zhang, Shen, Tong, and Wang}]{yang2025mmada}
Ling Yang, Ye~Tian, Bowen Li, Xinchen Zhang, Ke~Shen, Yunhai Tong, and Mengdi Wang. 2025{\natexlab{c}}.
\newblock Mmada: Multimodal large diffusion language models.
\newblock \emph{arXiv preprint arXiv:2505.15809}.

\bibitem[{Ye et~al.(2025)Ye, Xie, Zheng, Gao, Wu, Jiang, Li, and Kong}]{ye2025dream}
Jiacheng Ye, Zhihui Xie, Lin Zheng, Jiahui Gao, Zirui Wu, Xin Jiang, Zhenguo Li, and Lingpeng Kong. 2025.
\newblock Dream 7b: Diffusion large language models.
\newblock \emph{arXiv preprint arXiv:2508.15487}.

\bibitem[{Ye et~al.(2023)Ye, Zheng, Bao, Qian, and Gu}]{ye2023diffusion}
Jiasheng Ye, Zaixiang Zheng, Yu~Bao, Lihua Qian, and Quanquan Gu. 2023.
\newblock Diffusion language models can perform many tasks with scaling and instruction-finetuning.
\newblock \emph{arXiv preprint arXiv:2308.12219}.

\bibitem[{Yu et~al.(2025)Yu, Zhang, Zhu, Yuan, Zuo, Yue, Dai, Fan, Liu, Liu et~al.}]{yu2025dapo}
Qiying Yu, Zheng Zhang, Ruofei Zhu, Yufeng Yuan, Xiaochen Zuo, Yu~Yue, Weinan Dai, Tiantian Fan, Gaohong Liu, Lingjun Liu, and 1 others. 2025.
\newblock Dapo: An open-source llm reinforcement learning system at scale.
\newblock \emph{arXiv preprint arXiv:2503.14476}.

\bibitem[{Zhang et~al.(2024)Zhang, Wang, Diao, Lin, Pan, Dong, Zhang, Molchanov, and Zhang}]{zhang2024entropy}
Hanning Zhang, Pengcheng Wang, Shizhe Diao, Yong Lin, Rui Pan, Hanze Dong, Dylan Zhang, Pavlo Molchanov, and Tong Zhang. 2024.
\newblock Entropy-regularized process reward model.
\newblock \emph{arXiv preprint arXiv:2412.11006}.

\bibitem[{Zhang et~al.(2025)Zhang, Fang, Duan, He, Pan, Xiao, Huang, Zhai, Hu, Yu et~al.}]{zhang2025survey}
Lingzhe Zhang, Liancheng Fang, Chiming Duan, Minghua He, Leyi Pan, Pei Xiao, Shiyu Huang, Yunpeng Zhai, Xuming Hu, Philip~S Yu, and 1 others. 2025.
\newblock A survey on parallel text generation: From parallel decoding to diffusion language models.
\newblock \emph{arXiv preprint arXiv:2508.08712}.

\bibitem[{Zhao et~al.(2025{\natexlab{a}})Zhao, Gupta, Zheng, and Grover}]{zhao2025d1}
Siyan Zhao, Devaansh Gupta, Qinqing Zheng, and Aditya Grover. 2025{\natexlab{a}}.
\newblock d1: Scaling reasoning in diffusion large language models via reinforcement learning.
\newblock \emph{arXiv preprint arXiv:2504.12216}.

\bibitem[{Zhao et~al.(2025{\natexlab{b}})Zhao, Liu, Huang, Liu, Wang, Liu, Tian, Pang, Bell, Grover et~al.}]{zhao2025inpainting}
Siyan Zhao, Mengchen Liu, Jing Huang, Miao Liu, Chenyu Wang, Bo~Liu, Yuandong Tian, Guan Pang, Sean Bell, Aditya Grover, and 1 others. 2025{\natexlab{b}}.
\newblock Inpainting-guided policy optimization for diffusion large language models.
\newblock \emph{arXiv preprint arXiv:2509.10396}.

\bibitem[{Zheng et~al.(2025)Zheng, Liu, Li, Chen, Yu, Gao, Dang, Liu, Men, Yang et~al.}]{zheng2025group}
Chujie Zheng, Shixuan Liu, Mingze Li, Xiong-Hui Chen, Bowen Yu, Chang Gao, Kai Dang, Yuqiong Liu, Rui Men, An~Yang, and 1 others. 2025.
\newblock Group sequence policy optimization.
\newblock \emph{arXiv preprint arXiv:2507.18071}.

\bibitem[{Zhu et~al.(2025{\natexlab{a}})Zhu, Wang, Nie, Zhang, Wu, Hu, Zhou, Chen, Lin, Wen et~al.}]{zhu2025llada1_5}
Fengqi Zhu, Rongzhen Wang, Shen Nie, Xiaolu Zhang, Chunwei Wu, Jun Hu, Jun Zhou, Jianfei Chen, Yankai Lin, Ji-Rong Wen, and 1 others. 2025{\natexlab{a}}.
\newblock Llada 1.5: Variance-reduced preference optimization for large language diffusion models.
\newblock \emph{arXiv preprint arXiv:2505.19223}.

\bibitem[{Zhu et~al.(2025{\natexlab{b}})Zhu, You, Xing, Huang, Liu, Zhuang, Lu, Wang, Wang, Wei et~al.}]{zhu2025lladamoe}
Fengqi Zhu, Zebin You, Yipeng Xing, Zenan Huang, Lin Liu, Yihong Zhuang, Guoshan Lu, Kangyu Wang, Xudong Wang, Lanning Wei, and 1 others. 2025{\natexlab{b}}.
\newblock Llada-moe: A sparse moe diffusion language model.
\newblock \emph{arXiv preprint arXiv:2509.24389}.

\bibitem[{Zhu et~al.(2025{\natexlab{c}})Zhu, You, Xing, Huang, Liu, Zhuang, Lu, Wang, Wang, Wei et~al.}]{zhu2025llada}
Fengqi Zhu, Zebin You, Yipeng Xing, Zenan Huang, Lin Liu, Yihong Zhuang, Guoshan Lu, Kangyu Wang, Xudong Wang, Lanning Wei, and 1 others. 2025{\natexlab{c}}.
\newblock Llada-moe: A sparse moe diffusion language model.
\newblock \emph{arXiv preprint arXiv:2509.24389}.

\end{thebibliography}

\newpage
\appendix

\section{Proof of Theorem~1: High-probability Estimation Error Bound}
\label{app:proof}

\paragraph{Setup and Notation.} We consider one rollout \emph{tree step} in the tree-structured generation. Let $p$ denote the \emph{parent state}, i.e., the prompt together with a partially masked completion. In this tree step, the diffusion decoder reveals $k$ masked positions with index set
$\mathcal D=\{d_1,\dots,d_k\}$, producing the realized tokens $o^{d_1},\dots,o^{d_k}$.

\vspace{2pt}

\noindent A \emph{revelation order} is a permutation
\(
\sigma=(\sigma_1,\dots,\sigma_k)\in\Omega,
\)
where $\Omega$ is the set of all permutations of $\mathcal D$.
The decoding strategy induces a distribution $\mathcal Q$ over $\Omega$, i.e.,
\(
\sigma\sim \mathcal Q(\sigma).
\)
At sub-step $j$, the position revealed is $\sigma_j\in\mathcal D$.

\vspace{2pt}

\noindent For a given order $\sigma$, define the \emph{path-wise probability} at sub-step $j$ ($j\in [1,k]$) as
\begin{equation}
\small
\label{eq:app_qt}
q_j(\sigma)
=
f_\theta^{\sigma_j}\!\Big(
o^{\sigma_j}\mid p,\{o^{\sigma_1},\dots,o^{\sigma_{j-1}}\},\texttt{[MASK]}_{\text{else}}
\Big),
\end{equation}
where $\{o^{\sigma_1},\dots,o^{\sigma_{j-1}}\}$ denotes the already revealed realized tokens inserted back into the
current completion state, and all other unrevealed positions remain \texttt{[MASK]}.

\noindent \paragraph{Path-wise Probability.} For any position $d\in\mathcal D$ and order $\sigma\in\Omega$, let $\tau(d,\sigma)\in\{1,\dots,k\}$ be the unique index
such that $\sigma_{\tau(d,\sigma)}=d$. Thus, the realized token at position $d$ is revealed at sub-step $\tau(d,\sigma)$,
and the corresponding path-wise probability is $q_{\tau(d,\sigma)}(\sigma)$.

\noindent \paragraph{Single-time Forward Pass Estimate.} The single-time forward pass estimate is
\begin{equation}
\label{eq:app_phat}
\small
\hat p_d:= f_\theta^{d}(o^{d}\mid p).
\end{equation}

\noindent \paragraph{Unbiased Expected Probability.} We also recall the unbiased probability (expectation over all decoding orders) for completeness:
\begin{equation}
\label{eq:app_ptrue}
\small
p_d^{\mathrm{true}}
:=
\mathbb E_{\sigma\sim\mathcal Q}\big[q_{\tau(d,\sigma)}(\sigma)\big].
\end{equation}
Note that Theorem~1 bounds the deviation between $\hat p_d$ and the random path-wise probability
$q_{\tau(d,\sigma)}(\sigma)$ with high probability over $\sigma$, which in turn provides an explanation for why
$\hat p_d$ can serve as a reliable proxy when estimating the expectation $p_d^{\mathrm{true}}$.

\noindent\paragraph{Quantile-based confidence gap.} Fix a confidence level $\delta\in(0,1)$. For each decoded position $d\in\mathcal D$, define the
$(1-\delta)$-quantile of the random variable $q_{\tau(d,\sigma)}(\sigma)$ under $\sigma\sim\mathcal Q$ as
$q_{d,1-\delta}$, namely
\begin{equation}
\label{eq:app_quantile}
\small
\Pr_{\sigma\sim\mathcal Q}\!\big(q_{\tau(d,\sigma)}(\sigma)\ge q_{d,1-\delta}\big)\ge 1-\delta.
\end{equation}
We then define the (typical-path) confidence gap at level $\delta$:
\begin{equation}
\label{eq:app_eps}
\small
\epsilon_{d,\delta}:=\max\{1-\hat p_d,\ 1-q_{d,1-\delta}\}.
\end{equation}
By definition, $\hat p_d\ge 1-\epsilon_{d,\delta}$, and also $q_{d,1-\delta}\ge 1-\epsilon_{d,\delta}$.
Combining the latter with~\eqref{eq:app_quantile} yields the key high-probability event:
\begin{equation}
\small
\label{eq:app_good_event}
\Pr_{\sigma\sim\mathcal Q}\!\Big(q_{\tau(d,\sigma)}(\sigma)\ge 1-\epsilon_{d,\delta}\Big)\ge 1-\delta.
\end{equation}

\paragraph{Proof of Theorem~1.}
Fix any decoded position $d\in\mathcal D$ and assume $\epsilon_{d,\delta}\in[0,1)$.
Consider the event
\begin{equation}
\label{eq:app_E}
\small
E
:=
\Big\{\sigma\in\Omega:\ q_{\tau(d,\sigma)}(\sigma)\ge 1-\epsilon_{d,\delta}\Big\}.
\end{equation}
From~\eqref{eq:app_good_event}, we have $\Pr_{\sigma\sim\mathcal Q}(E)\ge 1-\delta$.

\vspace{2pt}

\noindent On the event $E$, we have the following two inequalities:
\begin{align}
q_{\tau(d,\sigma)}(\sigma) &\ge 1-\epsilon_{d,\delta}, \label{eq:app_q_lower}\\
\hat p_d &\ge 1-\epsilon_{d,\delta}. \label{eq:app_phat_lower}
\end{align}
Therefore, for any $\sigma\in E$,
\begin{equation}
\small
\label{eq:app_ratio_bounds}
1-\epsilon_{d,\delta}
\le
\frac{q_{\tau(d,\sigma)}(\sigma)}{\hat p_d}
\le
\frac{1}{1-\epsilon_{d,\delta}}.
\end{equation}
Indeed, the left inequality follows from $q_{\tau(d,\sigma)}(\sigma)\ge 1-\epsilon_{d,\delta}$ and $\hat p_d\le 1$;
the right inequality follows from $q_{\tau(d,\sigma)}(\sigma)\le 1$ and $\hat p_d\ge 1-\epsilon_{d,\delta}$.

\vspace{2pt}

\noindent Taking logarithms in~\eqref{eq:app_ratio_bounds} gives, for all $\sigma\in E$,
\begin{equation}
\small
\label{eq:app_log_bounds}
\log(1-\epsilon_{d,\delta})
\le
\log\frac{q_{\tau(d,\sigma)}(\sigma)}{\hat p_d}
\le
-\log(1-\epsilon_{d,\delta}),
\end{equation}
which implies
\begin{equation}
\small
\label{eq:app_abs_log}
\left|\log\frac{q_{\tau(d,\sigma)}(\sigma)}{\hat p_d}\right|
\le
-\log(1-\epsilon_{d,\delta}).
\end{equation}
Combining~\eqref{eq:app_abs_log} with $\Pr_{\sigma\sim\mathcal Q}(E)\ge 1-\delta$ proves
\begin{equation}
\small
\Pr_{\sigma\sim\mathcal Q}\!\left(
\left|\log\frac{q_{\tau(d,\sigma)}(\sigma)}{\hat p_d}\right|
\le
-\log(1-\epsilon_{d,\delta})
\right)\ge 1-\delta,
\end{equation}
which is exactly Eq.~\eqref{eq:theory} in the main text.

\paragraph{Additional Remarks.}
\begin{enumerate}
\item \textit{Why ``high probability''?}
The randomness comes from $\sigma\sim\mathcal Q$. Theorem~1 states that the estimation error is bounded on an event
of probability at least $1-\delta$ w.r.t.\ d$\mathcal Q$.
\item \textit{Why does confidence tighten the bound?}
The bound is monotone in $\epsilon_{d,\delta}$: as $\epsilon_{d,\delta}$ decreases,
$-\log(1-\epsilon_{d,\delta})$ decreases and approaches $0$. Thus, higher token confidence (smaller typical-path
confidence gap) yields a tighter error bound.
\item \textit{Relation to $p_d^{\mathrm{true}}$.}
Theorem~1 bounds the deviation between $\hat p_d$ and the random variable $q_{\tau(d,\sigma)}(\sigma)$ for typical
decoding orders. Since $p_d^{\mathrm{true}}$ is the expectation of $q_{\tau(d,\sigma)}(\sigma)$ over $\sigma\sim\mathcal Q$,
reducing the typical-path deviation (smaller $\epsilon_{d,\delta}$) makes $\hat p_d$ a more reliable proxy when
approximating $p_d^{\mathrm{true}}$.
\end{enumerate}

\section{Detailed Descriptions of dLLM RL Baselines}
\label{app:baselines}
\paragraph{Diffu-GRPO~\citep{zhao2025d1}.}
Diffu-GRPO adapts GRPO-style critic-free policy optimization to masked dLLMs by using sparse,
outcome-based rewards $r_i$ for each sampled completion $o_i$, and then assigning a group-relative advantage uniformly to all
tokens in that completion:
$A_{i,k}=r_i-\mathrm{mean}(\{r_j\}_{j=1}^G)$.
The key technical choice is its efficient log-probability approximation for dLLMs: instead of
marginalizing over denoising orders, it estimates per-token log-probabilities with a single
forward pass conditioned on a randomly masked prompt $q'$ and a fully masked completion, i.e.,
$\hat{\log \pi}_\theta(o_i^k \mid q')=\log f_\theta^k(o_i^k \mid q')$.
During optimization, Diffu-GRPO plugs these estimates into the standard GRPO clipped importance
ratio and KL-regularized objective, using $\hat{\pi}$ (denoted $\phi$ in their paper) for
$\pi_\theta,\pi_{\theta_{\mathrm{old}}},\pi_{\mathrm{ref}}$, and re-samples $q'$ each inner update as a
form of regularization while keeping $q,o_i,r_i$ fixed.

\paragraph{VRPO~\citep{zhu2025llada1_5}.} VRPO applies DPO-style preference optimization to masked dLLMs using offline preference pairs
$(q,o^{w},o^{l})$, where the training signal encourages the preferred completion $o^{w}$ to have a
higher relative score than $o^{l}$ without learning an explicit reward model. Since exact
$\log \pi_\theta(o\mid q)$ is intractable for dLLMs, VRPO estimates sequence log-probabilities
via an \emph{ELBO-based} surrogate $B_{\pi}(o\mid q)\le \log \pi(o\mid q)$ and substitutes it into the
DPO score:
\begin{equation}
\small
\begin{aligned}
s_\theta(o^{w},o^{l})
&=\beta\Big(B_{\pi_\theta}(o^{w}\!\mid q)-B_{\pi_{\mathrm{ref}}}(o^{w}\!\mid q)\Big) \\
&\quad-\beta\Big(B_{\pi_\theta}(o^{l}\!\mid q)-B_{\pi_{\mathrm{ref}}}(o^{l}\!\mid q)\Big).
\end{aligned}
\end{equation}
In practice, each $B_{\pi}(o\mid q)$ is approximated by a \emph{doubly} Monte Carlo estimator with
$n_t$ samples of the diffusion noise level (timestep) and $n_{z_t}$ masked-state samples per
timestep, which can induce substantial variance in the resulting preference score. The main
contribution of VRPO is to reduce this variance---most notably via an \emph{optimal allocation} of a
fixed sampling budget across timesteps (favoring larger $n_t$ and smaller $n_{z_t}$), and via
antithetic sampling that shares Monte Carlo randomness between the current policy and the reference
when forming the score differences. A remaining limitation is that, even with variance reduction, the
method does not explicitly account for the gap between the ELBO and the unbiased expected log-likelihood, since
the ELBO is only a lower bound. In their experiments, VRPO is applied to LLaDA-8B-Instruct,
producing the released model LLaDA-1.5.

\paragraph{wd1~\citep{tang2025wd1}.} wd1 is a diffusion-LM RL method that avoids GRPO/PPO-style \emph{policy ratios} and instead optimizes
a \emph{weighted log-likelihood} objective. For each prompt $q$, it samples a group of completions
$\{o_i\}_{i=1}^G$, computes a outcome reward $R(q,o_i)$ from a task-specific
verifier, and forms \emph{group-relative advantages}
$A_i = R(q,o_i) - \mathrm{mean}(R(q,o_{1:G}))$. Since the reward is only available at the completion
level, the same scalar signal is effectively broadcast to all tokens in $o_i$ via the sample weight, by
mapping $\{A_i\}$ to normalized weights
\begin{equation}
\small
w_i^{+}=\frac{\exp(\psi A_i)}{\sum_{j=1}^G \exp(\psi A_j)},
w_i^{-}=\frac{\exp(-\psi A_i)}{\sum_{j=1}^G \exp(-\psi A_j)},
\end{equation}
and optimizing
\begin{equation}
\small
\mathcal{L}_{\mathrm{wd1}}(\theta)
=\mathbb{E}\Big[\sum_{i=1}^G\big(-w_i^{+}+w_i^{-}\big)\,\log \pi_\theta(o_i\mid q)\Big].
\end{equation}
For probability estimation, wd1 still requires approximating the diffusion policy likelihood and adopts
the single-pass mean-field estimator,
$\log \pi_\theta(o_i\mid q)\approx \sum_k \log \pi_\theta(o_i^k\mid q')$,
where $q'$ is obtained by randomly masking the prompt at each gradient step.

\paragraph{SAPO~\citep{xie2025step}.} SAPO extends diffu-GRPO by augmenting the outcome reward with a process (step-aware) reward that
estimates whether an intermediate denoising interval makes progress toward a correct final answer.
For each prompt $q$, it samples a group of completions $\{o_i\}_{i=1}^G$ and computes scalar outcome
rewards $r_i$, then forms the group-relative advantage
$A_i = r_i - \mathrm{mean}(\{r_j\}_{j=1}^G)$. It further estimates a step-aware reward
$R_{\mathrm{process}}$ by comparing the expected outcome reward of rollouts continued from two
intermediate diffusion states; in practice it often uses the efficient case $t_2=T$ (all \texttt{[MASK]}),
reusing the original $G$ rollouts for the baseline term and sampling $N$ additional rollouts from a
randomly chosen intermediate state for the other term. SAPO combines these signals via
\begin{equation}
\small
    A^{\mathrm{total}}_i \;=\; A_i \;+\; \mathbf{1}[A_i>0]\cdot R_{\mathrm{process}} .
\end{equation}
During optimization, SAPO directly replaces the GRPO advantage with $A^{\mathrm{total}}_i$ in the same
PPO/GRPO-style clipped objective, i.e., the scalar $A^{\mathrm{total}}_i$ is \emph{broadcast} to all tokens
of $o_i$ (no token/step-specific advantage), and is used with token-wise likelihood ratios and a KL
regularizer. For probability estimation, SAPO inherits diffu-GRPO’s masked-dLLM likelihood
approximation, using mean-field per-token log-probabilities under randomly masked conditioning to
form ratios between the current and old policies (and the KL term to a reference).

\paragraph{GDPO~\citep{rojas2025improving}.} GDPO is a GRPO-style verifier-RL algorithm for diffusion LMs that keeps the same
\emph{sequence-level} outcome reward $R(q,o_i)$ and
group-relative advantage $A_i = R_i-\mathrm{mean}(R_{1:G})$,
but changes the \emph{probability estimation} and the \emph{importance ratio} from token-level
(mean-field) to \emph{sequence-level} via an ELBO surrogate. Concretely, GDPO defines a
sequence-level ratio using ELBO estimates
\begin{equation}
\small
    r_i(q,o_i) \;=\; \frac{L_{\mathrm{ELBO}}(o_i\mid q;\theta)}{L_{\mathrm{ELBO}}(o_i\mid q;\theta_{\mathrm{old}})},
\end{equation}
and plugs $r_g$ and $A_g$ into the same PPO/GRPO-style clipped objective (with a KL penalty to
$\pi_{\mathrm{ref}}$); thus the scalar advantage is still broadcast at the sample level rather than giving
token-/step-specific credit. The key technical contribution is a lower-variance, budget-efficient ELBO
estimator: it replaces the outer Monte Carlo over diffusion time $t$ with a small set of deterministic
quadrature points $\{(t_n,w_n)\}_{n=1}^N$ (typically $N=2$ or $3$) and uses a lightweight inner Monte
Carlo over masked states, yielding a semi-deterministic approximation
$L_{\mathrm{ELBO}}(o\mid q)\approx \sum_{n=1}^N w_n\,\ell(\pi_\theta; o,q,t_n)$.
Notably, like other ELBO-based approaches, GDPO optimizes and forms ratios using this lower bound and does not explicitly
account for the gap relative to the unbiased expected probability.

\paragraph{TraceRL~\citep{wang2025revolutionizing}.} TraceRL is a trajectory-aware PPO-style framework for dLLMs that optimizes on the model’s
actual decoding trajectory rather than random masking. For each prompt $q$, a rollout yields a
trajectory of intermediate states $(z_1,\ldots,z_0)$ with $z_0=o$ (the final completion), and the sets
of tokens revealed at each step (optionally aggregated by a shrinkage factor $s$ for efficiency). It
starts from a verifiable outcome reward $r(z_0)$ and can provide \emph{fine-grained} credit assignment
by training a diffusion-based value model $V_{\theta_v}$ and constructing token-/step-wise advantages
$A_j$ (e.g., GAE-style) along $(z_1,\ldots,z_0)$ instead of broadcasting a single scalar advantage to
all tokens; however, because these advantages depend on a learned value model, they bring
the usual risk of value mis-specification and reward/value hacking relative to purely verifier-based
outcome rewards. For probability estimation, TraceRL computes PPO importance ratios \emph{along the
realized trajectory}: for a token $o_j$ decoded when transitioning from $z_t$ to $z_{t-1}$, it uses
\begin{equation}
\small
r_j \;=\; \frac{\pi_{\theta}(o_j \mid z_t)}{\pi_{\mathrm{old}}(o_j \mid z_t)}.
\end{equation}
A remaining gap is that these likelihood terms are defined conditional on the single sampled
trajectory $(z_1,\ldots,z_0)$, whereas the true $\pi(o\mid q)$ would require marginalizing over all
possible decoding trajectories leading to $z_0=o$; TraceRL does not explicitly quantify or correct
this trajectory-marginalization gap.
\section{Training Details}
\label{app:training-details}
\subsection{Prompts}
Prompts used in Sudoku, Countdown, GSM8k and Math500 are as follows:
\begin{MyBox}
### Sudoku

Please solve the following 4x4 Sudoku puzzle. The puzzle is provided as a 16-character string reading left-to-right, top-to-bottom, where '0' represents empty cells.

Rules:
- Fill empty cells with digits 1-4
- Each row must contain digits 1-4 exactly once
- Each column must contain digits 1-4 exactly once
- Each 2x2 box must contain digits 1-4 exactly once

Important: Your solution must be a COMPLETE 16-character string with only the digits 1-4, representing your final solved grid.

Respond in this exact format:
<reasoning>
Your step-by-step solving process
</reasoning>
<answer>
[16-character solution string with no spaces or separators]
</answer>

Solve the following Sudoku puzzle: {PUZZLE}
\end{MyBox}
\begin{MyBox}
### Countdown

Respond in the following format:
<reasoning>
...
</reasoning>
<answer>
...
</answer>

Using only the numbers {NUMBERS}, create an arithmetic expression that evaluates to exactly {TARGET}. You must use all numbers from the list, and each number must be used exactly once. You may use the operations +, -, *, and / as needed. After reasoning, provide only your final expression inside <answer></answer> tags without including an equals sign or the target number. For example, if the numbers are [2, 3, 4] and the target is 5, a valid answer is: <answer>2*4-3</answer>.
\end{MyBox}
\begin{MyBox}
### GSM8k

Respond in the following format:
<reasoning>
...
</reasoning>
<answer>
...
</answer>

You are a math expert. You will be given a question to solve. Solve it step by step. Put only the final answer (the number) in the <answer> </answer> tag without any other text.

{QUESTION}
\end{MyBox}
\begin{MyBox}
### Math500

Respond in the following format:
<reasoning>
...
</reasoning>
<answer>
...
</answer>

You are a math expert. You will be given a question to solve. Solve it step by step. Wrap the final answer in a \\boxed{{}}.

{QUESTION}
\end{MyBox}

\subsection{Reward Functions}
\noindent\textbf{Sudoku.} The reward is defined as the percentage of correctly filled previously empty cells, with a maximum value of 1 and a minimum value of 0.

\noindent\textbf{Countdown.} The reward is 1.0 if the provided expression achieves the target using the given numbers. If the numbers used are correct but the target is missed, the reward is 0.1. In all other cases, the reward is 0.

\noindent\textbf{GSM8k.} During training, we utilize a correctness-based reward, assigning a value of 1.0 if the provided answer matches the ground truth.

\noindent\textbf{Math500.} Similarly, during training, we employ a correctness-based reward system, assigning a reward of 1.0 if the given answer matches the ground truth.

\subsection{Dataset Links}
Following previous works~\citep{zhao2025d1,tang2025wd1}, we adopt the 4×4 Sudoku training and test sets provided by Diffu-GRPO~\citep{zhao2025d1} for Sudoku. For Countdown, models are trained on the 3to4 Countdown task training set\footnote{\url{https://huggingface.co/datasets/Jiayi-Pan/Countdown-Tasks-3to4}} and evaluated using the test set provided by Diffu-GRPO. For GSM8K\footnote{\url{https://huggingface.co/datasets/openai/gsm8k}} and Math500\footnote{train: \url{https://huggingface.co/datasets/ankner/math-500}, test: \url{https://huggingface.co/datasets/HuggingFaceH4/MATH-500}}, we strictly follow their official training-test splits for both training and evaluation.

\section{Training Dynamics}
\subsection{Reward Curves Compared with Open-source Baselines}
\label{app:dynamics-1}
\begin{figure*}[t]
    \centering
    \includegraphics[width=\linewidth]{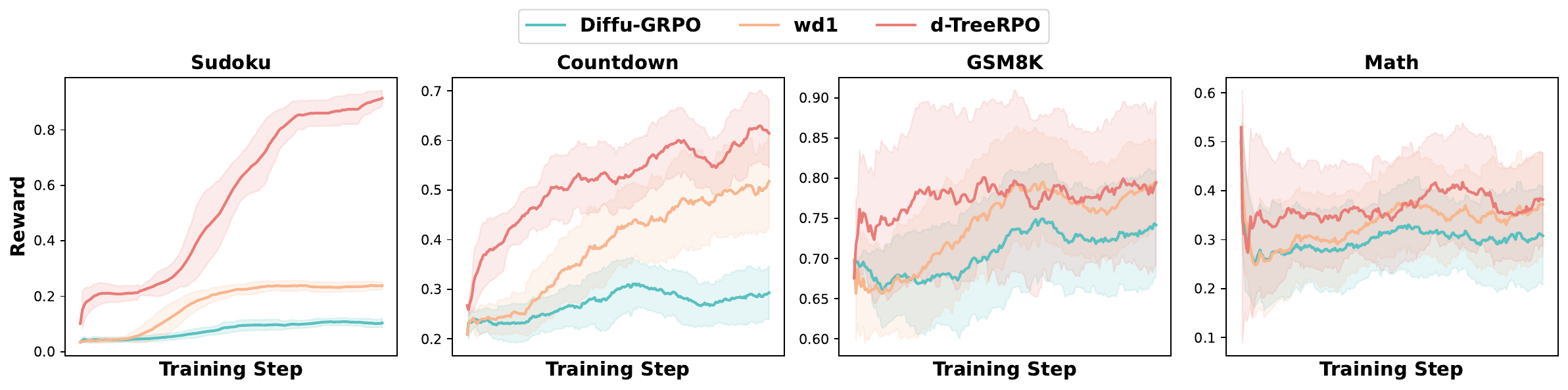}
    \caption{Comparison of training reward curves between \emph{d}-TreeRPO and open-source baselines: Diffu-GRPO~\citep{zhao2025d1} and wd1~\citep{tang2025wd1}.}
    \label{fig:rewards}
\end{figure*}

Figure \ref{fig:rewards} compares the training reward curves of \emph{d}-TreeRPO against two open-source baselines: Diffu-GRPO~\citep{zhao2025d1} and wd1~\citep{tang2025wd1}. We include these baselines because both our training code and wd1's implementation are built upon the Diffu-GRPO framework, ensuring better alignment of training parameters and making the curve comparison more meaningful. The figure specifically compares the training rewards of these three methods after processing \textit{equivalent amounts of data} (i.e., at the same point on the x-axis) across different tasks. It demonstrates that \emph{d}-TreeRPO consistently outperforms the baselines across all tasks, with particularly significant advantages in planning tasks like Sudoku and Countdown. wd1 achieves the second-highest performance, followed by Diffu-GRPO.
\label{app:dynamics}


\subsection{More Training Dynamics Regarding Self-distillation Loss}
\label{app:dynamics-2}
\begin{figure*}[h!]
    \centering 

    \begin{subfigure}[t]{0.242\textwidth}
        \centering
        \includegraphics[width=\textwidth]{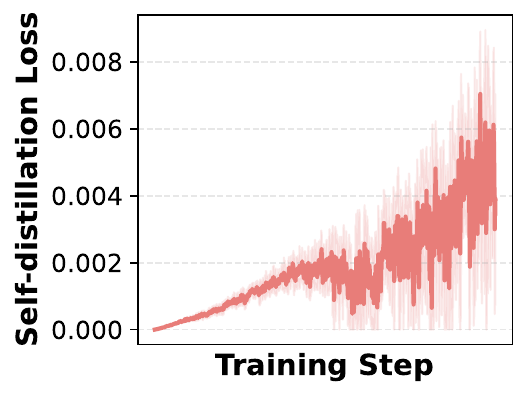} 
        \caption{Sudoku}
        \label{fig:label1}
    \end{subfigure}
    \hfill 
    \begin{subfigure}[t]{0.242\textwidth}
        \centering
        \includegraphics[width=\textwidth]{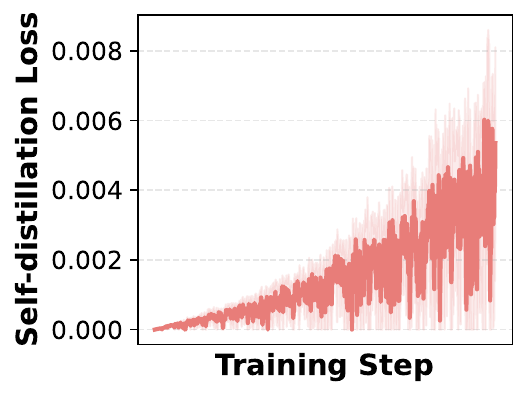} 
        \caption{Countdown}
        \label{fig:label2}
    \end{subfigure}
    \hfill 
    \begin{subfigure}[t]{0.242\textwidth}
        \centering
        \includegraphics[width=\textwidth]{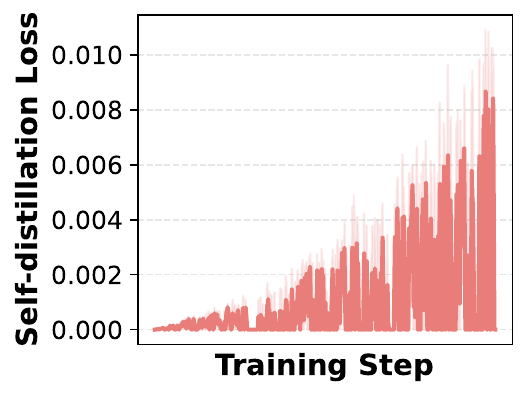} 
        \caption{GSM8k}
        \label{fig:label3}
    \end{subfigure}
    \begin{subfigure}[t]{0.242\textwidth}
        \centering
        \includegraphics[width=\textwidth]{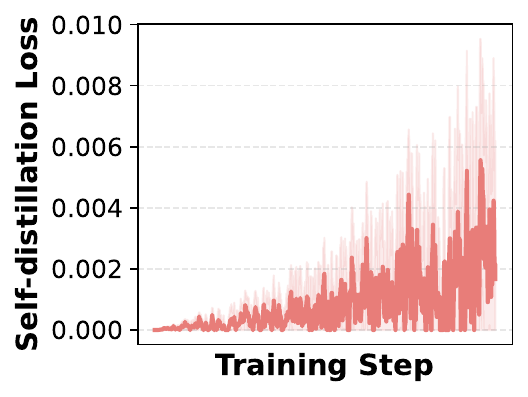} 
        \caption{Math500}
        \label{fig:label4}
    \end{subfigure}
    
    \caption{Self-distillation loss with $\lambda(t)$ over the course of training on the four evaluated tasks.}
    \label{fig:self-distillation-loss-with-λ}
\end{figure*}

\begin{figure*}[h!]
    \centering 

    \begin{subfigure}[t]{0.242\textwidth}
        \centering
        \includegraphics[width=\textwidth]{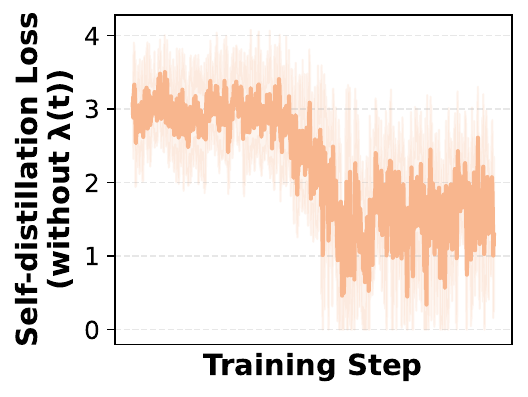} 
        \caption{Sudoku}
        \label{fig:label1}
    \end{subfigure}
    \hfill 
    \begin{subfigure}[t]{0.242\textwidth}
        \centering
        \includegraphics[width=\textwidth]{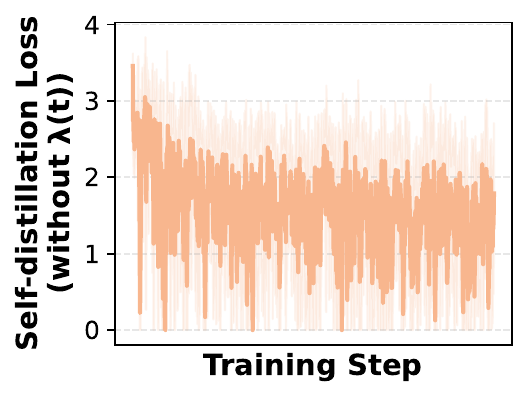} 
        \caption{Countdown}
        \label{fig:label2}
    \end{subfigure}
    \hfill 
    \begin{subfigure}[t]{0.242\textwidth}
        \centering
        \includegraphics[width=\textwidth]{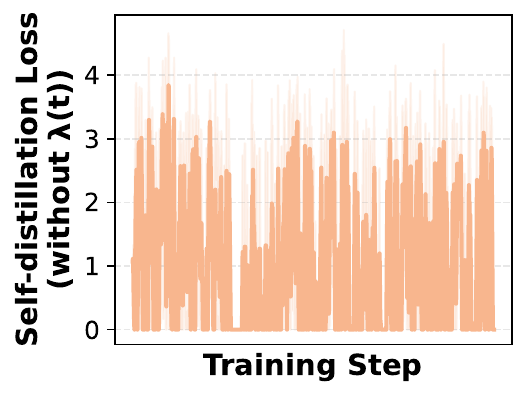} 
        \caption{GSM8k}
        \label{fig:label3}
    \end{subfigure}
    \begin{subfigure}[t]{0.242\textwidth}
        \centering
        \includegraphics[width=\textwidth]{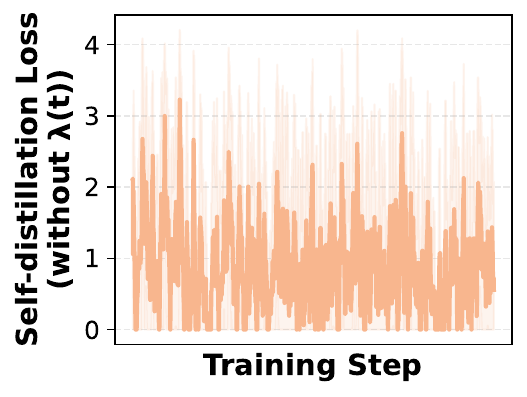} 
        \caption{Math500}
        \label{fig:label4}
    \end{subfigure}
    
    \caption{Self-distillation loss without $\lambda(t)$ over the course of training on the four evaluated tasks.}
    \label{fig:self-distillation-loss-without-λ}
\end{figure*}

\begin{figure*}[t]
    \centering 

    \begin{subfigure}[t]{0.242\textwidth}
        \centering
        \includegraphics[width=\textwidth]{fig/sudoku_entropy_curves.pdf} 
        \caption{Sudoku}
        \label{fig:label1}
    \end{subfigure}
    \hfill 
    \begin{subfigure}[t]{0.242\textwidth}
        \centering
        \includegraphics[width=\textwidth]{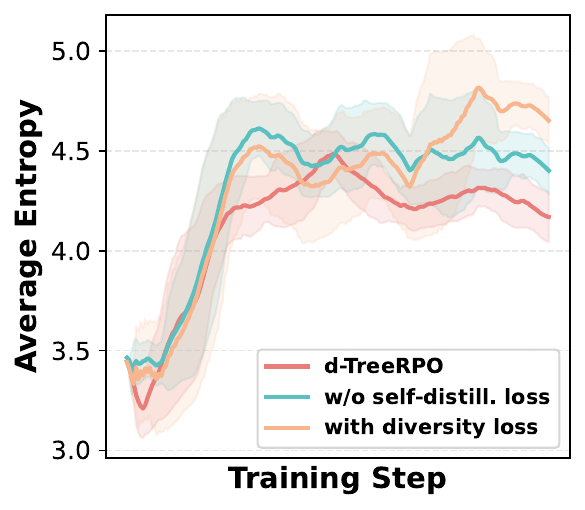} 
        \caption{Countdown}
        \label{fig:label2}
    \end{subfigure}
    \hfill 
    \begin{subfigure}[t]{0.242\textwidth}
        \centering
        \includegraphics[width=\textwidth]{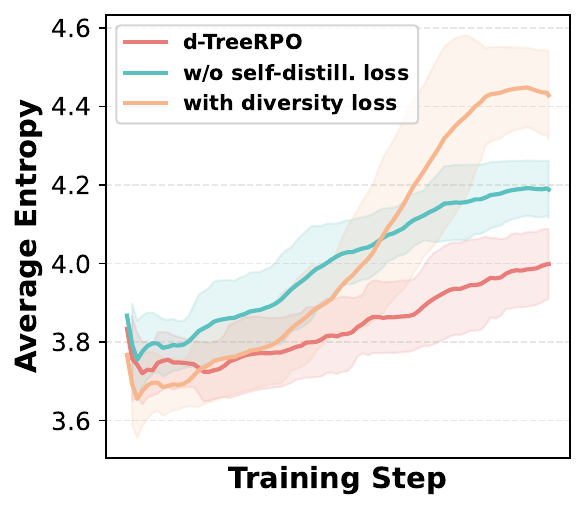} 
        \caption{GSM8k}
        \label{fig:label3}
    \end{subfigure}
    \begin{subfigure}[t]{0.242\textwidth}
        \centering
        \includegraphics[width=\textwidth]{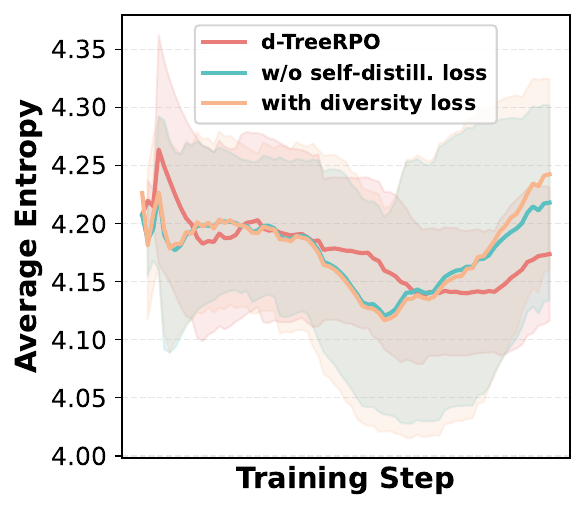} 
        \caption{Math500}
        \label{fig:label4}
    \end{subfigure}
    
    \caption{Entropy curves over the course of training on the four evaluated tasks under three settings: the full \emph{d}-TreeRPO algorithm, a variant without the self-distillation loss, and a variant with the diversity-promoting loss.}
    \label{fig:entropy-curves-all}
\end{figure*}

\begin{figure*}[t]
    \centering 

    \begin{subfigure}[t]{0.242\textwidth}
        \centering
        \includegraphics[width=\textwidth]{fig/sudoku_rewards.pdf} 
        \caption{Sudoku}
        \label{fig:label1}
    \end{subfigure}
    \hfill 
    \begin{subfigure}[t]{0.242\textwidth}
        \centering
        \includegraphics[width=\textwidth]{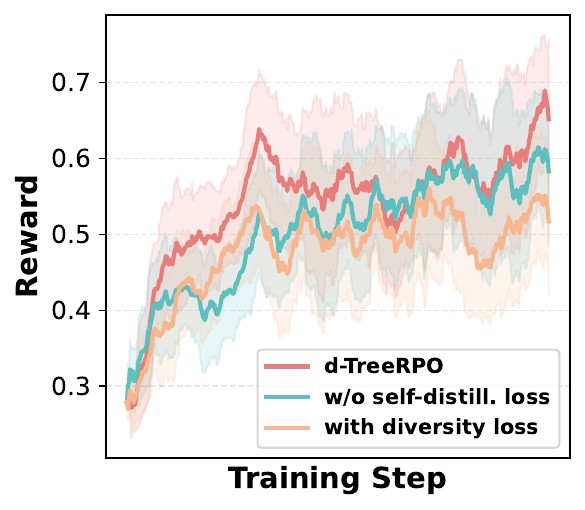} 
        \caption{Countdown}
        \label{fig:label2}
    \end{subfigure}
    \hfill 
    \begin{subfigure}[t]{0.242\textwidth}
        \centering
        \includegraphics[width=\textwidth]{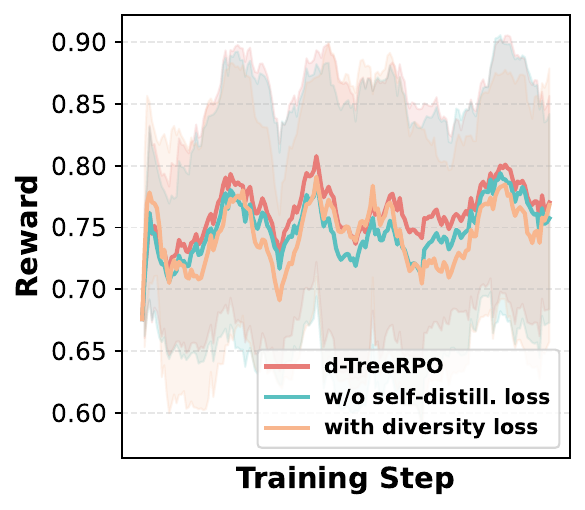} 
        \caption{GSM8k}
        \label{fig:label3}
    \end{subfigure}
    \begin{subfigure}[t]{0.242\textwidth}
        \centering
        \includegraphics[width=\textwidth]{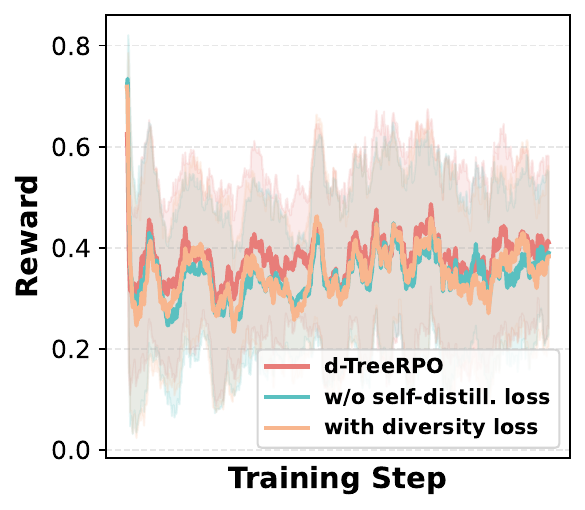} 
        \caption{Math500}
        \label{fig:label4}
    \end{subfigure}
    
    \caption{Reward curves over the course of training on the four evaluated tasks under three settings: the full \emph{d}-TreeRPO algorithm, a variant without the self-distillation loss, and a variant with the diversity-promoting loss.}
    \label{fig:reward-curves-all}
\end{figure*}

In this section, we present the evolution of the self-distillation loss over the course of training on all evaluated tasks, both with and without the weighting parameter $\lambda(t)$ (Figures~\ref{fig:self-distillation-loss-with-λ} and~\ref{fig:self-distillation-loss-without-λ}, respectively). We also report the entropy curves (Figure~\ref{fig:entropy-curves-all}) and training reward curves (Figure~\ref{fig:reward-curves-all}) under three settings: the full \emph{d}-TreeRPO algorithm, a variant without the self-distillation loss, and a variant with the diversity-promoting loss. These results serve as supplementary experiments to Section~\ref{sec:exp-effective} in the main text.

As shown in Figure~\ref{fig:self-distillation-loss-with-λ}, the self-distillation loss on all four tasks gradually increases as training progresses, which aligns with our design motivation: we impose a relatively loose constraint early on to encourage exploration, and then gradually tighten it later to promote determinism. This reduces probability estimation error and leads to better convergence performance. If we instead examine the self-distillation loss without multiplying by $\lambda(t)$, Figure~\ref{fig:self-distillation-loss-without-λ} shows an overall decreasing trend over training, which is particularly pronounced on planning tasks such as Sudoku and Countdown. This is because, as the self-distillation objective is strengthened, the policy progressively move closer to the predictions with positive advantages, causing the loss to decrease.

To further confirm the role of the self-distillation loss, we extend the experiments in Section~\ref{sec:exp-effective} to all four datasets. The entropy and reward curves are shown in Figures~\ref{fig:entropy-curves-all} and~\ref{fig:reward-curves-all}, respectively. We observe that, across all four datasets, the full \emph{d}-TreeRPO algorithm with the self-distillation loss achieves the lowest entropy in the later stage of training, followed by the variant without the self-distillation loss, while the variant with the diversity-promoting loss exhibits the highest entropy. The reward curves show a consistent pattern at convergence: the full \emph{d}-TreeRPO achieves the highest reward, the variant without self-distillation is second, and the diversity-promoting variant performs the worst. Since the overall performance gains on GSM8K and Math500 are relatively small for all methods, the benefit of adding the self-distillation loss is less pronounced, though still observable.

\section{Hyper-Parameter Analysis}
\label{app:hyper}

In this section, we provide additional hyper-parameter analyses for \emph{d}-TreeRPO by reporting evaluation performance on the four datasets under different hyper-parameter settings.

\subsection{Hyper-parameters regarding target distribution in self-distillation loss: $\tau_{max}$ and $\beta$}

Recall that the self-distillation loss uses a KL constraint to pull the current policy toward a target distribution obtained by voting among positive-advantage samples. This target distribution is advantage-weighted and defined as:
\begin{equation}
\small
P_{\text{target}}^{\sigma_i}(v) =
\frac{\sum_{c\in C_p^+} w_c \cdot \mathbf{1}[v_c^{\sigma_i} = v]}{\sum_{c\in C_p^+} w_c},
\quad \forall v \in \mathcal{V}.
\end{equation}

The weight $w_c$ is computed by applying a temperature-scaled softmax to the advantages:
\begin{equation}
\small
w_c =
\frac{\exp\left(A_p^c/\tau(t)\right)}{\sum_{c'\in C_p^+} \exp\left(A_p^{c'}/\tau(t)\right)},
\quad \forall c \in C_p^+.
\end{equation}

A smaller $\tau(t)$ makes the target distribution more sharply concentrated on the branch with the largest positive advantage, whereas a larger $\tau(t)$ yields a more uniform weighting across all positive-advantage branches. In our implementation, we control $\tau(t)$ using a time-dependent schedule:
\begin{equation}
\small
\tau(t) = \tau_{\max} \cdot \left(1 - \frac{t}{T}\right)^\beta, \quad (\beta \in (0,1]),
\end{equation}
where $T$ denotes the total number of training steps and $t$ is the current training step.

\begin{figure*}[h!]
    \centering 

    \begin{subfigure}[t]{0.242\textwidth}
        \centering
        \includegraphics[width=\textwidth]{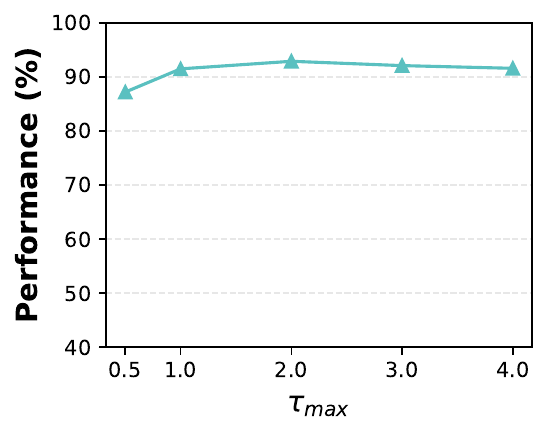} 
        \caption{Sudoku}
        \label{fig:label1}
    \end{subfigure}
    \hfill 
    \begin{subfigure}[t]{0.242\textwidth}
        \centering
        \includegraphics[width=\textwidth]{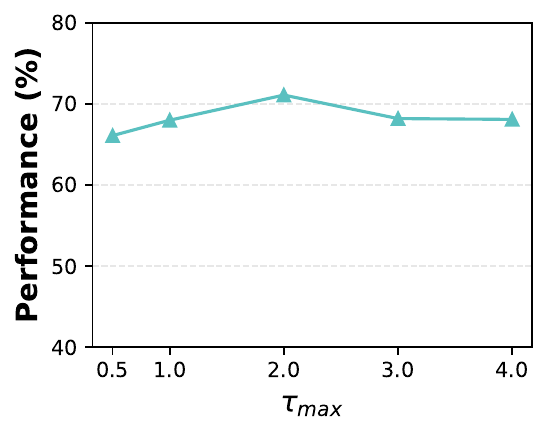} 
        \caption{Countdown}
        \label{fig:label2}
    \end{subfigure}
    \hfill 
    \begin{subfigure}[t]{0.242\textwidth}
        \centering
        \includegraphics[width=\textwidth]{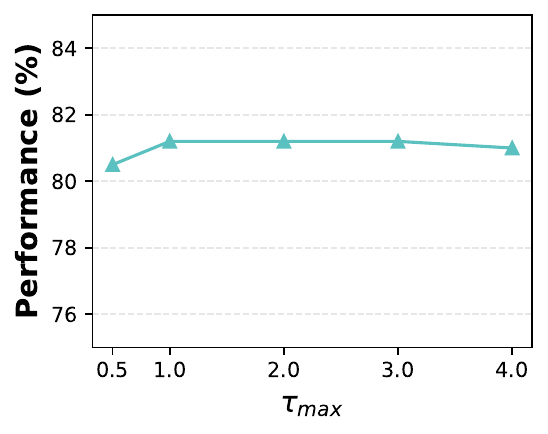} 
        \caption{GSM8k}
        \label{fig:label3}
    \end{subfigure}
    \begin{subfigure}[t]{0.242\textwidth}
        \centering
        \includegraphics[width=\textwidth]{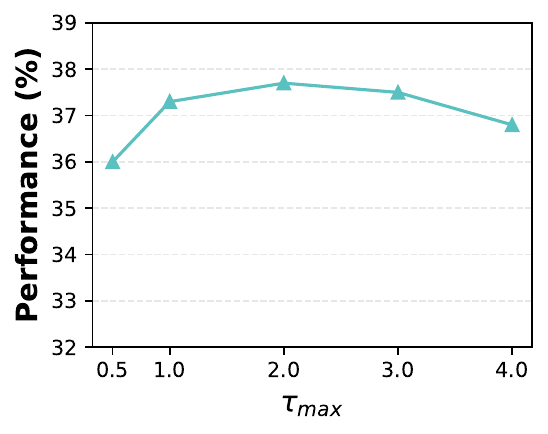} 
        \caption{Math500}
        \label{fig:label4}
    \end{subfigure}
    
    \caption{Evaluation performance under different $\tau_{max}$ settings on the four evaluated tasks.}
    \label{fig:hyper-tau}
\end{figure*}

This schedule introduces two hyper-parameters, $\tau_{\max}$ and $\beta$. A larger $\tau_{\max}$ corresponds to a higher overall temperature and, consequently, higher entropy. As shown in Figure~\ref{fig:hyper-tau}, \emph{d}-TreeRPO is generally robust to the choice of $\tau_{\max}$. Upon closer inspection, overly large or overly small values of $\tau_{\max}$ can slightly degrade performance, either by reducing determinism or by making the policy overly deterministic (thereby hindering exploration). Therefore, we set $\tau_{\max}=2.0$ in our main experiments.

\begin{figure*}[h!]
    \centering 

    \begin{subfigure}[t]{0.242\textwidth}
        \centering
        \includegraphics[width=\textwidth]{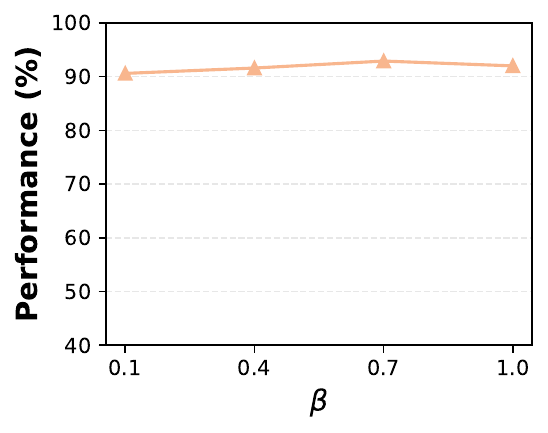} 
        \caption{Sudoku}
        \label{fig:label1}
    \end{subfigure}
    \hfill 
    \begin{subfigure}[t]{0.242\textwidth}
        \centering
        \includegraphics[width=\textwidth]{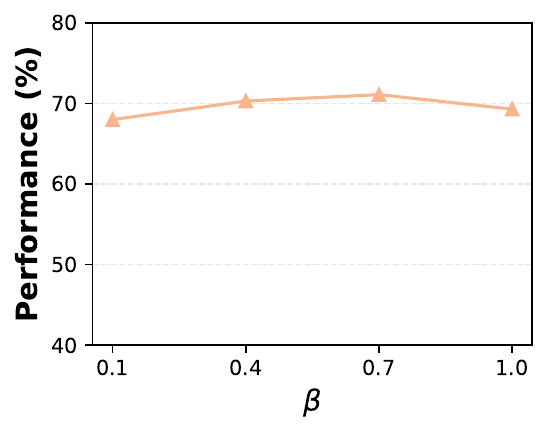} 
        \caption{Countdown}
        \label{fig:label2}
    \end{subfigure}
    \hfill 
    \begin{subfigure}[t]{0.242\textwidth}
        \centering
        \includegraphics[width=\textwidth]{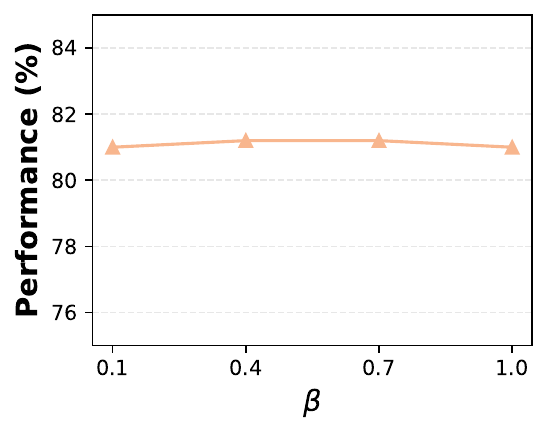} 
        \caption{GSM8k}
        \label{fig:label3}
    \end{subfigure}
    \begin{subfigure}[t]{0.242\textwidth}
        \centering
        \includegraphics[width=\textwidth]{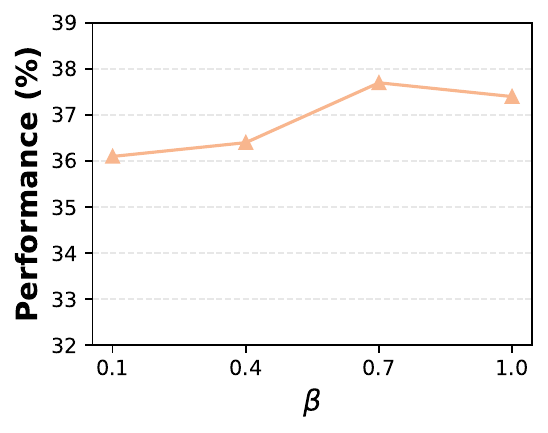} 
        \caption{Math500}
        \label{fig:label4}
    \end{subfigure}
    
    \caption{Evaluation performance under different $\beta$ settings on the four evaluated tasks.}
    \label{fig:hyper-beta}
\end{figure*}

The exponent $\beta \in (0,1]$ controls the decay rate: smaller $\beta$ leads to a slower decrease in $\tau(t)$ early in training and a much faster drop later. When $\beta$ is close to $0$, the entropy-reducing effect of the self-distillation loss becomes weaker, while $\beta$ close to $1$ yields an approximately linear schedule. As shown in Figure~\ref{fig:hyper-beta}, \emph{d}-TreeRPO is also fairly robust to the choice of $\beta$. When $\beta$ approaches $0$, the behavior becomes close to constructing the self-distillation target distribution using uniform weights over positive samples, which slightly degrades performance. When $\beta$ approaches $1$, the performance changes only marginally. Based on the performance trends, we set $\beta=0.7$ in our main experiments.

\begin{figure*}[h!]
    \centering 

    \begin{subfigure}[t]{0.242\textwidth}
        \centering
        \includegraphics[width=\textwidth]{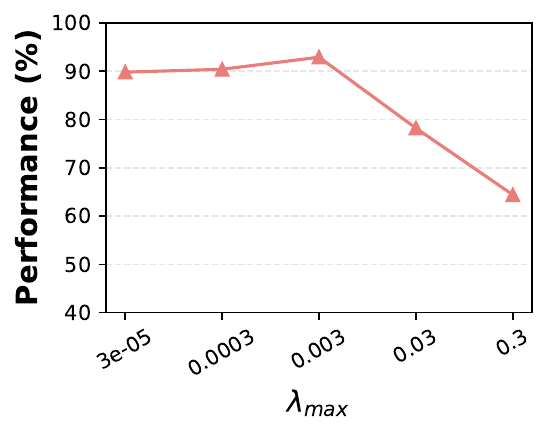} 
        \caption{Sudoku}
        \label{fig:label1}
    \end{subfigure}
    \hfill 
    \begin{subfigure}[t]{0.242\textwidth}
        \centering
        \includegraphics[width=\textwidth]{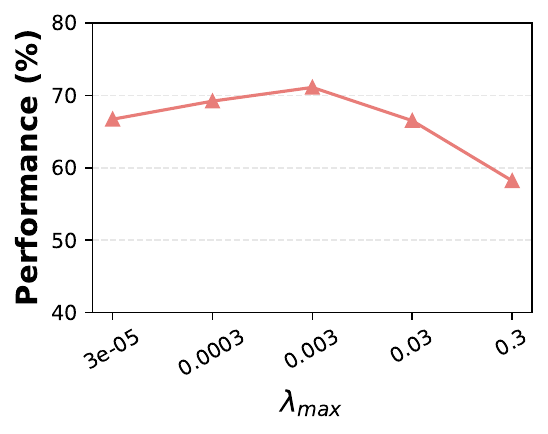} 
        \caption{Countdown}
        \label{fig:label2}
    \end{subfigure}
    \hfill 
    \begin{subfigure}[t]{0.242\textwidth}
        \centering
        \includegraphics[width=\textwidth]{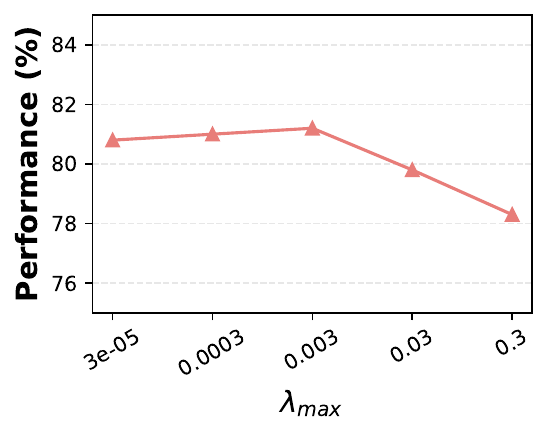} 
        \caption{GSM8k}
        \label{fig:label3}
    \end{subfigure}
    \begin{subfigure}[t]{0.242\textwidth}
        \centering
        \includegraphics[width=\textwidth]{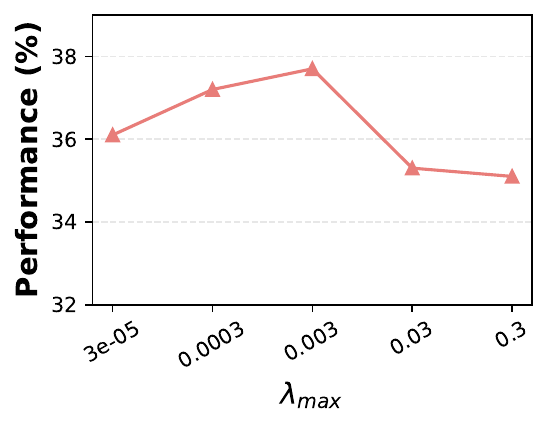} 
        \caption{Math500}
        \label{fig:label4}
    \end{subfigure}
    
    \caption{Evaluation performance under different $\lambda_{max}$ settings on the four evaluated tasks.}
    \label{fig:hyper-lambda}
\end{figure*}

\begin{figure*}[h!]
    \centering 

    \begin{subfigure}[t]{0.242\textwidth}
        \centering
        \includegraphics[width=\textwidth]{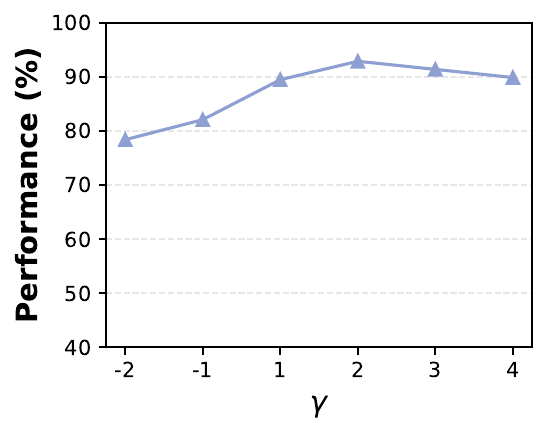} 
        \caption{Sudoku}
        \label{fig:label1}
    \end{subfigure}
    \hfill 
    \begin{subfigure}[t]{0.242\textwidth}
        \centering
        \includegraphics[width=\textwidth]{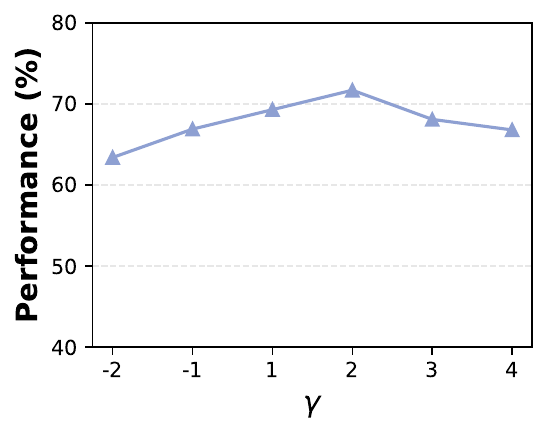} 
        \caption{Countdown}
        \label{fig:label2}
    \end{subfigure}
    \hfill 
    \begin{subfigure}[t]{0.242\textwidth}
        \centering
        \includegraphics[width=\textwidth]{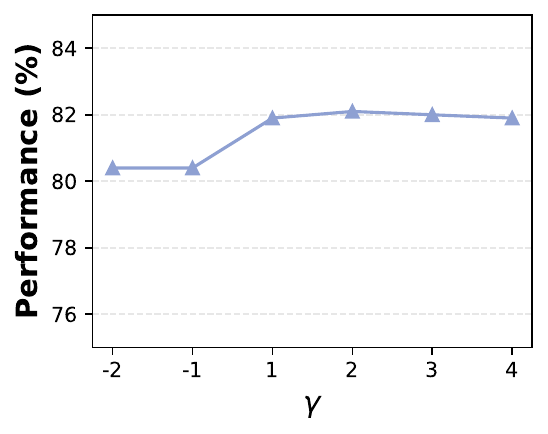} 
        \caption{GSM8k}
        \label{fig:label3}
    \end{subfigure}
    \begin{subfigure}[t]{0.242\textwidth}
        \centering
        \includegraphics[width=\textwidth]{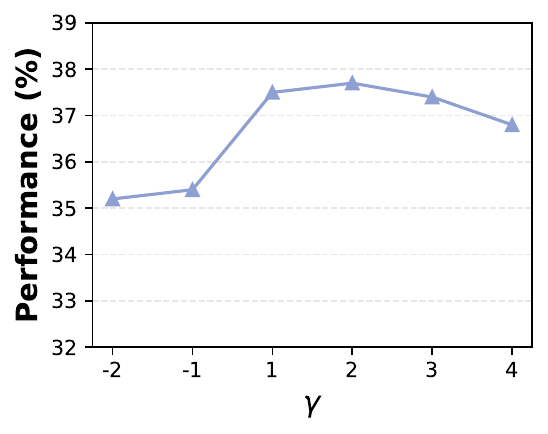} 
        \caption{Math500}
        \label{fig:label4}
    \end{subfigure}
    
    \caption{Evaluation performance under different $\gamma$ settings on the four evaluated tasks.}
    \label{fig:hyper-gamma}
\end{figure*}

\subsection{Hyper-parameters regarding $\lambda(t)$: $\gamma$ and $\lambda_{max}$}

In this subsection, we focus on how the coefficient $\lambda(t)$ in the self-distillation loss affects performance. Recall that $\lambda(t)$ is defined as
\begin{equation}
    \lambda(t) = \lambda_{\text{max}} \cdot \frac{e^{\gamma t/T}-1}{e^\gamma - 1},
\label{eq:lambda}
\end{equation}
which introduces two hyper-parameters: $\lambda_{\max}$ and $\gamma$. Larger $\lambda_{\max}$ increases the overall scale of the self-distillation loss and thus strengthens its effect in promoting determinism. As shown in Figure~\ref{fig:hyper-lambda}, when $\lambda_{\max}$ is too small, the behavior becomes similar to removing the self-distillation loss altogether. In contrast, when $\lambda_{\max}$ is too large---especially on the order of $0.3$---it can substantially hinder exploration and degrade performance. Therefore, to balance the exploration--exploitation trade-off, and based on the results in Figure~\ref{fig:hyper-lambda}, we set $\lambda_{\max}=3\times 10^{-3}$ in our main experiments.

The parameter $\gamma$ controls how fast $\lambda(t)$ increases. When $\gamma>0$, $\lambda(t)$ grows slowly at the beginning and accelerates later; when $\gamma<0$, it exhibits the opposite behavior. As shown in Figure~\ref{fig:hyper-gamma}, settings with $\gamma>0$ generally outperform those with $\gamma<0$, because increasing the weight too quickly in the early stage can suppress exploration. A schedule with an increasing first derivative is therefore preferable. Based on these results, we choose $\gamma=2$ as the default setting in our main experiments.

\section{Case Studies}
This section presents case studies of \emph{d}-TreeRPO, highlighting its advantages over alternative training methods. We compare the original LLaDA-8B-Instruct with models trained using Diffu-GRPO, wd1, and \emph{d}-TreeRPO on the same GSM8K question, as shown in Figures~\ref{fig:llada-8b-instruct-case}, \ref{fig:d1-case}, \ref{fig:wd1-case}, and~\ref{fig:dtreerpo-case}, respectively. Among these methods, only \emph{d}-TreeRPO produces both a correct reasoning process and the correct final answer.
\label{app:case}
\begin{figure*}[t]
\begin{case}[title=LLaDA-8B-Instruct]
    \textbf{Question:} Ten treeks weigh as much as three squigs and one goolee. Two treeks and one goolee are equal in weight to one squig. The combined weight of how many treeks equals the weight of one squig?

    \vspace{3pt}
    
    \textbf{Model Output:} 
    
    \texttt{<reasoning>} 
    
    To solve this problem, we need to establish the relationships between the treeks, squigs, and goolees based on the given information. Let's denote the weight of one treek as $T$, the weight of one squig as $S$, and the weight of one goolee as $G$. 
    
    From the problem, we have the following relationships:
    
    1. Ten treeks weigh as much as three squigs and one goolee: $10T = 3S + G$
    
    2. Two treeks and one goolee are equal in weight to one squig: $2T + G = S$
    
    We need to find the number of treeks whose combined weight equals the weight of one squig. From the second relationship, we know that $2T + G = S$. This implies that the combined weight of two treeks and one goolee is equal to the weight of one squig. \textcolor{red}{Therefore, the weight of two treeks equals the weight of one squig. \textcolor{deepred}{\ding{55}}}
    
    \texttt{</reasoning>}
    
    \texttt{<answer>}
    
    $\boxed{2}$ \textcolor{deepred}{\ding{55}}
    
    \texttt{</answer>}
\end{case}
\caption{A case study of LLaDA-8B-Instruct's response to a GSM8K question.}
\label{fig:llada-8b-instruct-case}
\end{figure*}

\begin{figure*}[t]
\begin{case}[title=LLaDA-8B-Instruct + Diffu-GRPO]
\textbf{Question:} Ten treeks weigh as much as three squigs and one goolee. Two treeks and one goolee are equal in weight to one squig. The combined weight of how many treeks equals the weight of one squig?

\vspace{3pt}

\textbf{Model Output:} 

\texttt{<reasoning>} 

To solve this problem, we need to establish the relationships between the treeks, squigs, and goolees based on the given information. Let's denote the weight of one treek as $T$, the weight of one squig as $S$, and the weight of one goolee as $G$.

From the problem, we have the following relationships:

1. Ten treeks weigh as much as three squigs and one goolee: $10T = 3S + G$

2. Two treeks and one goolee are equal in weight to one squig: $2T + G = S$

We need to find the number of treeks whose combined weight equals the weight of one squig. From the second relationship, we know that $2T + G = S$. This means that the combined weight of two treeks and one goolee is equal to the weight of one squig. \textcolor{red}{Therefore, the number of treeks that equal the weight of one squig is two. \textcolor{deepred}{\ding{55}}}

\texttt{</reasoning>}
    
\texttt{<answer>}

$\boxed{2}$ \textcolor{deepred}{\ding{55}}

\texttt{</answer>}
\end{case}
\caption{A case study of LLaDA-8B-Instruct trained with Diffu-GRPO responding to a GSM8K question.}
\label{fig:d1-case}
\end{figure*}

\begin{figure*}[t]
\begin{case}[title=LLaDA-8B-Instruct + wd1]
\textbf{Question:} Ten treeks weigh as much as three squigs and one goolee. Two treeks and one goolee are equal in weight to one squig. The combined weight of how many treeks equals the weight of one squig?

\vspace{3pt}

\textbf{Model Output:} 

\texttt{<reasoning>} 

1. Let the weight of one treek be $T$, the weight of one squig be $S$, and the weight of one goolee be $G$

2. From the first equation: $10T = 3S + G$

3. From the second equation: $2T + G = S$

4. Substitute $S = 2T + G$ from the second equation into the first: $10T = 3(2T + G) + G$, \textcolor{red}{$10T = 6 +T + G$,  $4T = G$} \textcolor{deepred}{\ding{55}}

5. Now, $S =2T + 4T = 6T$

6. To find how many treeks equal the weight of one squig: $S = 6T$ \text{which means 6 treeks}

\texttt{</reasoning>} 

\texttt{<answer>}

$\boxed{6}$ \textcolor{deepred}{\ding{55}}

\texttt{</answer>}
\end{case}
\caption{A case study of LLaDA-8B-Instruct trained with wd1 responding to a GSM8K question.}
\label{fig:wd1-case}
\end{figure*}

\begin{figure*}[t]
\begin{case}[title=LLaDA-8B-Instruct + \emph{d}-TreeRPO]
\textbf{Question:} Ten treeks weigh as much as three squigs and one goolee. Two treeks and one goolee are equal in weight to one squig. The combined weight of how many treeks equals the weight of one squig?

\vspace{3pt}

\textbf{Model Output:} 

\texttt{<reasoning>} 

Let the weights in terms of a treek be $T$, a squig be $S$, and a goolee be $G$. From the problem statement, we have the following equations:

1. $10T = 3S + G$

2. $2T + G = S$

We need to find the number of treeks that equal the weight of one squig, i.e., $nT = S$.

First, let's solve the second equation for $G$: $G = S - 2T$. Substitute $G$ into the first equation: $10T = 3S + (S - 2T)$, $10T = 4S - 2T$, $10T + 2T = 4S$, $12T = 4S$, $3T = S$.

Thus, the weight of three treeks equals the weight of one squig.

\texttt{</reasoning>} 

\texttt{<answer>}

$\boxed{3}$ \textcolor{green!60!black}{\ding{51}}

\texttt{</answer>}
\end{case}
\caption{A case study of LLaDA-8B-Instruct trained with \emph{d}-TreeRPO responding to a GSM8K question.}
\label{fig:dtreerpo-case}
\end{figure*}

\end{document}